\documentclass{article}

\usepackage{microtype}
\usepackage{graphicx}
\usepackage{subfigure}
\usepackage{booktabs} 

\usepackage{hyperref}

\usepackage[accepted]{icml2025}

\usepackage{amsmath}
\usepackage{amssymb}
\usepackage{mathtools}
\usepackage{amsthm}

\usepackage[capitalize,noabbrev]{cleveref}

\usepackage{float}
\usepackage[retain-zero-uncertainty=true]{siunitx}
\usepackage[version=4]{mhchem}
\usepackage{multirow}

\theoremstyle{plain}

\theoremstyle{definition}

\theoremstyle{remark}

\usepackage[textsize=tiny]{todonotes}

\icmltitlerunning{Temperature-Annealed Boltzmann Generators}

\begin{document}

\twocolumn[
\icmltitle{Temperature-Annealed Boltzmann Generators}



\icmlsetsymbol{equal}{*}

\begin{icmlauthorlist}
\icmlauthor{Henrik Schopmans}{iti}
\icmlauthor{Pascal Friederich}{iti,int}
\end{icmlauthorlist}

\icmlaffiliation{int}{Institute of Nanotechnology, Karlsruhe Institute of Technology, Kaiserstr. 12, 76131 Karlsruhe, Germany}
\icmlaffiliation{iti}{Institute of Theoretical Informatics, Karlsruhe Institute of Technology, Kaiserstr. 12, 76131 Karlsruhe, Germany}

\icmlcorrespondingauthor{Pascal Friederich}{pascal.friederich@kit.edu}

\icmlkeywords{boltzmann generators, normalizing flows, molecular systems, variational sampling}

\vskip 0.3in
]



\printAffiliationsAndNotice{}  

\begin{abstract}
    Efficient sampling of unnormalized probability densities such as the
    Boltzmann distribution of molecular systems is a longstanding challenge.
    Next to conventional approaches like molecular dynamics or Markov chain
    Monte Carlo, variational approaches, such as training normalizing flows with
    the reverse Kullback-Leibler divergence, have been introduced. However, such
    methods are prone to mode collapse and often do not learn to sample the full
    configurational space. Here, we present temperature-annealed Boltzmann
    generators (TA-BG) to address this challenge. First, we demonstrate that
    training a normalizing flow with the reverse Kullback-Leibler divergence at
    high temperatures is possible without mode collapse. Furthermore, we
    introduce a reweighting-based training objective to anneal the 
    distribution to lower target temperatures.
    We apply this methodology to three molecular systems of increasing complexity 
    and, compared to the baseline, achieve better results in almost all metrics while requiring up to 
    three times fewer target energy evaluations. For the largest system, our approach is the 
    only method that accurately resolves the metastable states of the system.
    
\end{abstract}

\section{Introduction}
\label{introduction}
Machine learning, and particularly generative models, have become a
transformative force across numerous domains. A prime example of this impact is
in structural biology, where deep learning methods such as those in the
AlphaFold family
\cite{jumperHighlyAccurateProtein2021,abramsonAccurateStructurePrediction2024}
have revolutionized our ability to predict protein structures. While a big part
of AlphaFold's success can surely be attributed to an advanced methodology, a
key factor also lies in the availability of abundant experimental data, such as that in the Protein Data Bank (PDB)
\cite{burleyRCSBProteinData2021}.


However, not all scientific domains benefit from such well-curated and extensive
experimental datasets. In areas where data scarcity is a persistent challenge,
computational simulations play an essential role. Molecular dynamics (MD)
and Markov chain Monte Carlo (MCMC) methods are the primary tools used to
explore complex biochemical and physical systems and generate insights from
limited experimental information. Despite their utility, these classical
sampling approaches often come with significant computational costs, as they
rely on iterative trajectory-based exploration of high-dimensional state spaces.

As a result, various approaches have been explored to speed up these methods, including
integrating machine learning (ML)-based force fields
\cite{reiserGraphNeuralNetworks2022a}, enhanced sampling techniques
\cite{barducciMetadynamics2011}, and data-driven collective
variables \cite{bonatiDeepLearningSlow2021}. Furthermore, (transferable) generative models have
been trained on equilibrium samples from MD simulations
\cite{noeBoltzmannGeneratorsSampling2019a,mahmoudAccurateSamplingMacromolecular2022,zhengPredictingEquilibriumDistributions2024,kleinTransferableBoltzmannGenerators2024a}.

While these advancements have significantly improved the efficiency and utility
of traditional simulations, there is a growing interest in rethinking the
paradigm altogether. Variational sampling methods, rooted in generative
modeling, offer a compelling alternative to classical MD and MCMC. These
approaches aim to learn the underlying probability distribution without the availability of training data,
bypassing the need for explicit trajectory-based sampling.


The most straightforward variational approach is to train a likelihood-based
generative model, such as a normalizing flow, using the reverse Kullback-Leibler
divergence (KLD). However, this is known to yield mode collapse in many
scenarios
\cite{midgleyFlowAnnealedImportance2023,felardosDesigningLossesDatafree2023}.
Recently, multiple variational sampling methods were developed
\cite{blessingELBOsLargeScaleEvaluation2024}, based on normalizing flows
\cite{matthewsContinualRepeatedAnnealed2022,midgleyFlowAnnealedImportance2023},
diffusion models
\cite{zhangPathIntegralSampler2021,richterImprovedSamplingLearned2023,bernerOptimalControlPerspective2023,vargasTransportMeetsVariational2023,zhangDiffusionGenerativeFlow2023a,akhound-sadeghIteratedDenoisingEnergy2024a,senderaImprovedOffpolicyTraining2024},
and flow matching \cite{wooIteratedEnergybasedFlow2024}.

Despite their promise to accelerate sampling, the applicability and scalability
of variational sampling methods remain limited, and the field is in its early
stages of development compared to the wealth of research on hybrid MD/ML
approaches. To the best of our knowledge, the only variational approach that has
successfully been applied to the sampling of molecular systems with non-trivial
multimodality, such as the popular benchmark system alanine dipeptide, is Flow
Annealed Importance Sampling Bootstrap (FAB)
\cite{midgleyFlowAnnealedImportance2023}. 

In this work, we propose a novel and scalable flow-based framework to efficiently sample
complex molecular systems without mode collapse. We train a normalizing flow at
increased temperature using the reverse KLD, which we show reliably circumvents
mode collapse. Since one is typically interested in the equilibrium distribution
at lower temperatures, e.g. at room temperature, we introduce a reweighting-based training
objective to iteratively anneal the distribution of the normalizing flow down to
the target temperature. We demonstrate the capability and scalability of this methodology using three 
peptide systems of increasing complexity and achieve superior sampling efficiency and accuracy compared to baseline approaches.

Our contribution is threefold:
\begin{itemize}
    \item We show that, in contrast to current literature, the reverse KLD is
    surprisingly powerful at learning the Boltzmann distribution of molecular
    systems without mode collapse, but only at increased temperatures where barriers between different free energy minima are lower and the probability distribution maxima are interconnected.
    \item We introduce an iterative reweighting-based training strategy
    to anneal the flow distribution to arbitrary target temperatures.
    \item We introduce two complex molecular systems as new benchmarks that go far beyond the size of the typically used benchmark system alanine dipeptide, and we demonstrate that our approach scales to those systems without mode collapse.
\end{itemize}

\begin{figure*}[ht]
\vskip 0.2in
\begin{center}
\centerline{\includegraphics{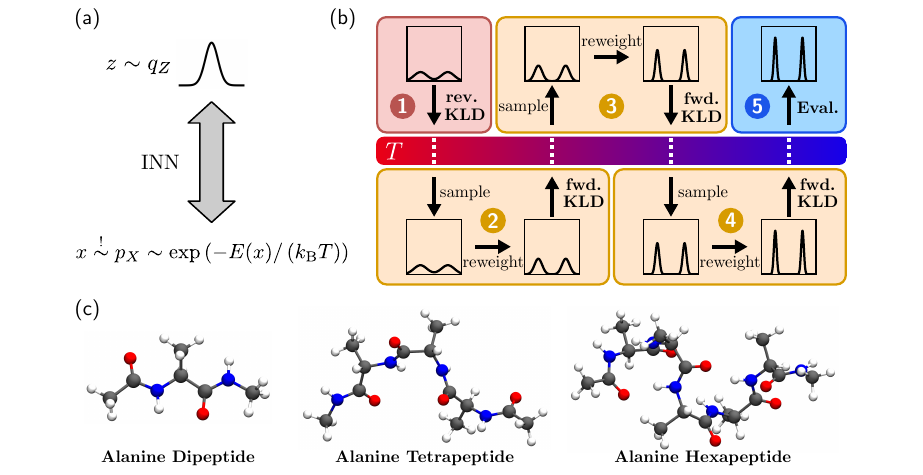}} \caption{(a)
Illustration of Boltzmann generators based on normalizing flows. The goal is to
learn the equilibrium Boltzmann distribution of the 3D conformations of a
molecular system. We focus on data-free training, where only the
unnormalized probability density is known. (b) Illustration of our workflow. (1) To
avoid mode collapse, we first train the flow at high temperature with the
reverse KLD. (2-4) Then, the learned distribution is iteratively annealed to the
target temperature by sampling at the current temperature, reweighting these
samples to a lower temperature, and subsequently performing forward KLD training
at this lower temperature. This is repeated multiple times until the desired
target temperature is reached. (5) In the end, samples are drawn from the flow
for evaluation. (c) Visualization of the three peptide molecular systems used in this work.}
\label{main:fig1}
\end{center}
\vskip -0.2in
\end{figure*}

\section{Related Work} \label{sec:related_work}
Leveraging the improved mode-mixing behavior when sampling at higher
temperatures is not a completely novel approach, as it was introduced as an accelerated sampling technique for MCMC and MD simulations before. Replica exchange Markov chain Monte Carlo (RE-MCMC) and molecular dynamics (REMD)
use multiple parallel trajectories (replicas) at different temperatures, while allowing 
repeated exchanges of configurations between the replicas. This essentially makes the
high-temperature simulations help the lower-temperature simulations in overcoming
slow energy barriers in the system.

\citeauthor{invernizziSkippingReplicaExchange2022}
\yrcite{invernizziSkippingReplicaExchange2022} present a variation of replica
exchange molecular dynamics using a normalizing flow that maps from the highest
temperature directly to the target temperature. This allows direct
exchanges and circumvents the need for intermediate replicas in the simulation.
However, this still requires performing MD simulations at the boundary temperatures 
and it is not clear how well the method scales to larger systems.

\citeauthor{dibakTemperatureSteerableFlows2022}
\yrcite{dibakTemperatureSteerableFlows2022} use a normalizing flow that is
trained on samples from high-temperature molecular dynamics simulations. Using
a special flow architecture, they show that the flow can be adapted to 
output low-temperature samples, even though it was only trained at the high
temperature. \citeauthor{draxlerUniversalityVolumePreservingCouplingBased2024}
\yrcite{draxlerUniversalityVolumePreservingCouplingBased2024} recently showed
that the volume-preserving coupling layers used in that work are not universal, making
the approach unsuitable for complex systems.

To solve this issue, \citeauthor{schebekEfficientMappingPhase2024a}
\yrcite{schebekEfficientMappingPhase2024a} propose to use a normalizing flow
with explicit conditioning on the temperature $T$ and pressure $P$. The prior of
the normalizing flow is formed by samples from an MD simulation at a 
reference thermodynamic state $(T_0,P_0)$. The normalizing flow is subsequently
trained to be able to sample across a range of thermodynamic states $(T,P)$
using the reverse KLD.

Similarly, \citeauthor{wahlTRADETransferDistributions2025}
\yrcite{wahlTRADETransferDistributions2025} train a temperature-conditioned flow
at increased temperature using samples from MD, and the correct temperature
scaling to lower temperatures is obtained by matching the gradient of the
unnormalized probability density of the flow with respect to the temperature to
the gradient of the known target energy function.

So far, all mentioned approaches use (a large amount of) MD samples at at least one temperature, and
transfer this to a different (lower) temperature. While a multitude of
variational sampling methods that do not rely on samples from MD have been
proposed, to the best of our knowledge, the only approach that has so
far been successfully applied to non-trivial molecular systems is Flow Annealed
Importance Sampling Bootstrap (FAB) by
\citeauthor{midgleyFlowAnnealedImportance2023}
\yrcite{midgleyFlowAnnealedImportance2023}. 

Instead of the reverse KLD, FAB uses the $\alpha$-divergence with $\alpha=2$ for
energy-based training.
The $\alpha$-divergence
is estimated using annealed importance sampling (AIS)
\cite{nealAnnealedImportanceSampling2001} from $q_X$ to $\frac{p_X^2}{q_X}$,
where Hamiltonian Monte Carlo (HMC) \cite{duaneHybridMonteCarlo1987} is used to
transition between intermediate distributions. While AIS sampling is costly in
terms of energy evaluations, it proved effective in learning complex probability
distributions. They successfully learned the Boltzmann distribution of alanine
dipeptide, a common benchmark molecular system, without mode collapse. However,
how well this method scales to more complex systems is unclear, and we will
address this in this work when comparing our approach to FAB.

\section{Preliminaries} \label{sec:preliminaries}

\subsection{Normalizing Flows}
Normalizing flows use a latent distribution $q_Z(z)$, typically
a Gaussian or uniform distribution, which is transformed to the target space
using an invertible transformation $ x = g(z;\theta) $, $z = f(x;\theta) =
g^{-1}(x;\theta)$ (Figure~\ref{main:fig1}a).

The transformed density of the flow can be expressed using the change of
variables formula \cite{dinhNICENonlinearIndependent2015}:

\begin{align}
    &q_X(x;\theta)=q_Z\left(f(x;\theta)\right)\left|\operatorname{det} J_{x \mapsto z}\right| \label{eq:norm_flow_transform} \\
    &\text{with the Jacobian } J_{x \mapsto z} = \frac{\partial f(x;\theta)}{\partial x^T} \notag
\end{align}

The most common approach to parameterize the invertible function is to use
invertible coupling layers. In each coupling layer, the input $x_{1:D}$ is split
into two parts $x_{1:d}$ and $x_{d+1:D}$. The first part is transformed
elementwise conditioned on the second part, while the second part is kept
identical (see Figure~\ref{SI:coupling_flow} in the appendix for an illustration). 
If the elementwise transformation is invertible (monotonic), the
whole transformation becomes invertible. Furthermore, the Jacobian matrix of
such a coupling transform is lower triangular and can be efficiently
computed \cite{durkanNeuralSplineFlows2019a}.

\paragraph{Training with Samples.}
The key property of normalizing flows is that the likelihood (Equation~\ref{eq:norm_flow_transform}) is directly available, typically at the cost of a forward pass. This allows data-based maximum likelihood training (forward KLD):

\begin{align}
    &\mathrm{KL}_{\theta}\left[p_X \| q_X\right] =C-\int p_X(x) \log q_X(x ; \theta) \, \mathrm{d} x \\
    & =C-\mathbb{E}_{x \sim p_X}{\left[\log q_Z\left(f(x;\theta)\right) + \log \left|\operatorname{det} J_{x \mapsto z}\right| \right]}
    \label{eq:flow_training_by_example}
\end{align}

\paragraph{Training by Energy.}
Next to data-based training, a normalizing flow can be trained by energy if only
the target density $p_X(x)$ is known. In case of physical systems, such as the
molecules studied in this work, this is the Boltzmann distribution $p_X(x) \sim
\exp \left( -E(x) / (k_{\mathrm{B}} T) \right)$. Here, $x$ is the 3D
configuration of the molecule, for example, the Cartesian coordinates of all
atoms, $k_{\mathrm{B}}$ is the Boltzmann constant, $T$ is the temperature, and
$E(x)$ is the energy of the given configuration, evaluated either using quantum
mechanics, e.g. with density functional theory, or, as in this work, using a
parameterized force field. While other data-free training objectives exist (such
as the $\alpha=2$ divergence used in FAB), the reverse KLD is the most
straightforward objective to fit the distribution of the flow to the target
density, using samples from the flow itself:

\begin{align}
    &\mathrm{KL}_{\theta}\left[q_X \| p_X\right] = \mathrm{KL}_{\theta}\left[q_Z \| p_Z\right] \\ 
    &= C-\int q_Z(z) \log p_Z(z ; \theta) \, \mathrm{d} z \\
    &= C-\mathbb{E}_{z \sim q_Z} \left[\log p_X\left(g(z;\theta)\right) + \log \left|\operatorname{det} J_{z \mapsto x}\right| \right]
    \label{eq:flow_training_by_energy}
\end{align}


\paragraph{Importance Sampling}
Since normalizing flows provide the likelihood of the generated samples, one can
perform importance sampling to the true distribution $p_X$ using the
importance weights $w(x)=\frac{p_X(x)}{q_X(x;\theta)}$. When estimating an expectation
value of an observable $h(x)$ using samples $x_n$ from the flow distribution
$q_X$, this offers asymptotically unbiased estimates
\cite{martinoEffectiveSampleSize2017,noeBoltzmannGeneratorsSampling2019a}:

\begin{align} 
     \sum_{n=1}^N \frac{w(x_n)}{\sum_{i=1}^N w(x_n)} h\left(x_n\right) \xrightarrow[N \to \infty]{} \int h(x) p_X(x) \mathrm{d}x \label{eq:importance_sampling}
\end{align}

While diffusion models and continuous normalizing flows can provide likelihoods
by computing divergences of the involved vector field, this is often
prohibitively expensive in practice, already for relatively small systems
\cite{kleinTransferableBoltzmannGenerators2024a}.

While Equation \ref{eq:importance_sampling} theoretically allows unbiased
estimates, this is limited in practice by the actual overlap between the flow
distribution $q_X$ and the target distribution $p_X$. A helpful measure, here,
is the effective sample size (ESS), defined as the number of independent samples
needed from the target distribution $p_X$ to achieve the same variance in
estimating expectation values as when using the flow distribution $q_X$
\cite{martinoEffectiveSampleSize2017}. The reverse ESS is an approximation of
the ESS, where samples from the flow are used (see
Section~\ref{SI:sec:metrics_details} in the appendix). The ESS is thus only
estimated within the support of the flow, meaning that a high ESS can still be
achieved if only parts of the true distribution are covered. Therefore, the
reverse ESS value needs to always be interpreted together with other metrics.

\begin{figure*}[ht]
\vskip 0.2in
\begin{center}
\centerline{\includegraphics{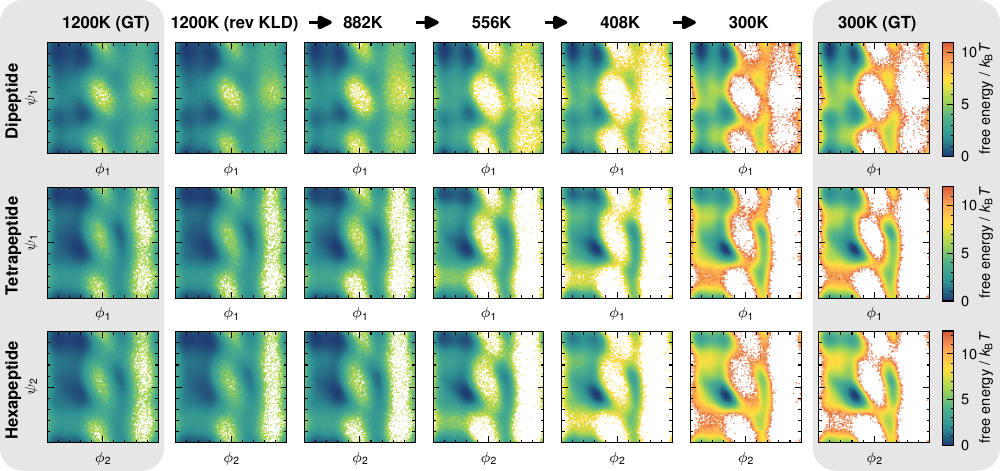}} \caption{Visualization
of the iterative annealing process, showing the free energy $ F=-k_\text{B} T
\ln p(\phi_i, \psi_i) $ of backbone dihedral angles (Ramachandran plots) in each
iteration. After learning the distribution at \SI{1200}{\kelvin} using the
reverse KLD, the distribution is annealed step by step to the target temperature
\SI{300}{\kelvin}. Note that not all annealing iterations are shown. Since the
tetrapeptide has three pairs of backbone dihedral angles, and the hexapeptide
five, we selected one pair of dihedral angles for this illustration. The
annealing of all pairs of backbone dihedrals can be found in
Figures~\ref{SI:tetra_annealing} and \ref{SI:hexa_annealing} in the appendix. We
used \num{1e7} samples for the Ramachandran plots at \SI{300}{\kelvin} and
\num{1e6} samples for the rest.}
\label{main:fig_annealing}
\end{center}
\vskip -0.2in
\end{figure*}

\section{Methods} \label{sec:methods} 
\subsection{Flow Architecture}
Analogous to previous works
\cite{midgleyFlowAnnealedImportance2023,schopmansConditionalNormalizingFlows2024}, we
use an internal coordinate representation based on bond lengths, angles,
and dihedral angles to represent the molecular conformations. This incorporates
the symmetries of the potential energy, which is invariant to translations and
rotations of the whole molecule.
For all experiments, we use a normalizing flow built from 16 monotonic rational-quadratic spline coupling layers \cite{durkanNeuralSplineFlows2019a} 
with fully connected parameter networks in the couplings. Dihedral angles are treated using circular splines \cite{rezendeNormalizingFlowsTori2020a}
to incorporate the correct topology.
Details can be found in Sections~\ref{SI:internal_coordinates} and \ref{SI:architecture} of the appendix.

\subsection{Temperature-Annealed Boltzmann Generators}
Our approach to learn the Boltzmann distribution of molecular systems can be
separated into two phases (see Figure~\ref{main:fig1}b): First, we learn the
distribution at a high temperature using the reverse KLD (step 1 in
Figure~\ref{main:fig1}b). Due to decreased barrier heights, this can be done
without mode collapse. Secondly, the distribution is iteratively annealed using
the mass-covering forward KLD with importance-sampled datasets to obtain the
distribution at the target temperature. We now explain both steps in detail and
discuss why they avoid the challenges described above.

\subsection*{Training by Energy: Avoiding Mode Collapse}
The mode-seeking behavior of the reverse KLD has been discussed and observed in 
multiple previous publications \cite{midgleyFlowAnnealedImportance2023,felardosDesigningLossesDatafree2023,soletskyiTheoreticalPerspectiveMode2024}.
Once the flow collapsed 
to a mode, meaning that some remaining modes of the target distribution are not within 
the support of the flow, it will generally not escape this collapsed state if the remaining 
modes are too far separated from the collapsed mode.
This is not surprising, since the reverse KLD is evaluated using an expectation value with samples 
from the flow distribution $q_X$ itself, 
which will only cover the collapsed modes.

At the typical target temperature for molecular systems, i.e. \SI{300}{\kelvin},
modes are too far separated to be successfully covered with the reverse KLD.
However, when sampling at increased temperature, the modes become more
connected. Eventually, one can use the reverse KLD to efficiently learn the
distribution. In this work, we performed the reverse KLD experiments at
\SI{1200}{\kelvin}, which allows a relatively small batch size and number of
gradient descent steps to be used without mode collapse (see
Section~\ref{SI:sec:ta_bg_ablations} in the appendix for an extended ablation
and discussion regarding choosing the starting temperature).

Using any loss function that directly includes the target energy of a molecular
system can be challenging. If two atoms overlap sufficiently, the repulsive van
der Waals energy diverges, leading to unstable training. Following previous work
\cite{midgleyFlowAnnealedImportance2023}, we thus use a regularized energy
function for training (see Section~\ref{sec:SI:mol_systems} in the appendix).

This avoids very high values in the loss function and stabilizes training.
Furthermore, analogous to previous work \cite{schopmansConditionalNormalizingFlows2024}, we found that removing a 
small fraction of the largest energy values in the loss contributions of each batch stabilizes training.

\subsection*{Reweighting-Based Annealing}
As explained, we use the reverse KLD as a first step to learn the Boltzmann
distribution at increased temperature. To obtain the distribution at a lower
target temperature, here \SI{300}{\kelvin}, we utilize importance sampling
(Equation~\ref{eq:importance_sampling}). While one could do importance sampling
directly from $T_\text{start}=\SI{1200}{\kelvin}$ to
$T_\text{target}=\SI{300}{\kelvin}$, this will yield bad overlap and sampling
efficiency (see Figure~\ref{SI:fig_direct_importance_sampling} in the appendix).

Instead, we perform importance sampling using multiple temperatures $ T_1 =
T_\text{start}, T_2, T_3, \ldots, T_{K-1}, T_K = T_\text{target}$, where $T_1
\ge T_2 \ge T_3 \ge \ldots \ge T_K$. In one annealing iteration, we perform the
following steps:

\begin{enumerate}
    \item Sample a dataset $\mathcal{D}_i = \{x_j\}_{j=1}^N$ of $N$ samples $x_j \sim q_X(x_j;\theta)$ from the flow at the current temperature $T_i$.
    \item Calculate importance weights $w(x_j) = \frac{p_{X, T=T_{i+1}}(x_j)}{q_X(x_j;\theta)}$ for each sample $x_j$ in $\mathcal{D}_i$ to transition to $T_{i+1}$.
    \item According to these importance weights, resample a dataset $\mathcal{W}_{i+1}$ with replacement from $\mathcal{D}_i$.
    \item Perform forward KLD training on $\mathcal{W}_{i+1}$.
\end{enumerate}

Throughout this annealing workflow, we keep updating the flow parameters, without
reinitialization. In this way, we can anneal the distribution of the flow step
by step toward the target temperature $T_\text{target}$ (see steps 2-4 in
Figure~\ref{main:fig1}b). Since we use the mass-covering forward KLD objective
based on importance-sampled datasets, mode collapse is not a problem during the
annealing.

For all experiments, we chose 9 temperature annealing iterations. To ensure a
similar overlap between two consecutive distributions, we choose the
temperatures $T_i$ using a geometric progression between $T_\text{start}$ and
$T_\text{target}$ \cite{sugitaReplicaexchangeMolecularDynamics1999} (see
Section~\ref{SI:sec:ta_bg_ablations} in the appendix for a comparison with a
linear temperature schedule):

\begin{align}
T_i = T_\text{start} \left( \frac{T_\text{target}}{T_\text{start}} \right)^{\frac{i-1}{K-1}}
\end{align}

Furthermore, we added a final fine-tuning iteration, where we sample at
\SI{300}{\kelvin} and reweight to \SI{300}{\kelvin}
($T_{K-1}=T_K=T_\text{target}$). Since the temperature is not lowered in this
fine-tuning step, the effective sample size of the training dataset is higher,
which empirically improves the final metrics obtained at \SI{300}{\kelvin}. For
the hexapeptide system, we added such a fine-tuning iteration with $T_{i+1}=T_i$
after each annealing iteration (see Equation~\ref{SI:eq:hexa_temp_schedule} in
the appendix). While this increases the total number of target potential energy
evaluations, it improves the obtained results substantially. For the two less
complex systems, intermediate fine-tuning was not necessary. We refer to
Section~\ref{SI:sec:ta_bg_ablations} of the appendix for an extended discussion
and ablations regarding the fine-tuning iterations.

\subsection*{Variations}
We note that our buffered reweighting approach is not the only option to anneal 
the temperature of a normalizing flow. As discussed in Section~\ref{sec:related_work}, a concurrent 
study \cite{wahlTRADETransferDistributions2025} uses a temperature-conditioned normalizing flow 
with a temperature scaling loss to learn the Boltzmann distribution at the target temperature.
While this approach achieves promising results, it requires the repeated estimation of the partition function $Z$ using importance sampling with the flow distribution.
Obtaining a low-variance estimate of $Z$ can be computationally expensive, especially for high-dimensional systems such as the hexapeptide studied here. However, a systematic comparison of different temperature scaling approaches is
an interesting avenue for future work.

We further experimented with variations of our reweighting approach. For example, we tried training a temperature-conditioned 
flow with a reweighting-based objective continuously on the whole temperature range. This is described in more detail in 
Section~\ref{SI:FAB_variations} of the appendix.
In practice, we found the iterative buffered annealing workflow to be superior, both in 
terms of accuracy and sampling efficiency.

\section{Experiments}
We now describe the conducted experiments to evaluate our temperature-annealing
approach. The objective is to learn the Boltzmann distribution of three
molecular systems, increasing in complexity (see Figure~\ref{main:fig1}c). The
first molecule is alanine dipeptide, a popular system that previously served as
a benchmark for variational sampling \cite{midgleyFlowAnnealedImportance2023},
but also other related tasks such as transition path sampling
\cite{holdijkStochasticOptimalControl2023,seongTransitionPathSampling2024}. We
further evaluate on two higher-dimensional and more complex systems, alanine
tetrapeptide and alanine hexapeptide. All three systems have complex metastable
high-energy regions that make up only a small fraction of the entire state
space, which makes them suitable hard objectives for benchmarking.

While the focus of our work is on molecular systems due to their challenging potential energy surface,
TA-BG can also be applied to other sampling tasks.
We refer to the appendix, Section~\ref{SI:sec_gmm}, for additional experiments on a Gaussian mixture system.

\subsection{Baseline Methods}
To judge the performance of our approach, we compare it to baseline methods. First, 
we trained a normalizing flow with the forward KLD using MD data
from the target distribution. While this is not a variational
sampling approach, it serves as a good baseline to show the expressiveness of
the flow if data is available. Next, we trained a normalizing flow with the
reverse KLD, targeting the Boltzmann distribution at \SI{300}{\kelvin}.
As the final and most powerful baseline, we trained Flow Annealed Importance
Sampling Bootstrap (FAB) \cite{midgleyFlowAnnealedImportance2023}. As already
discussed, to the best of our knowledge, this is the only method that so far has
shown success without mode collapse on our smallest test system alanine
dipeptide. It therefore serves as a strong baseline.

\subsection{Metrics}
To evaluate the distribution of the normalizing flow at the target temperature
\SI{300}{\kelvin}, we use a combination of multiple metrics. First, we use the
negative log-likelihood (NLL) calculated on the ground truth test dataset. This is a
good overall measure of the learned distribution. We note that since the
metastable regions of our systems form only a small part of the ground truth
test datasets, seemingly minor differences in the NLL can be decisive,
especially when assessing mode collapse or the quality of the description of the
metastable region.

To better assess potential mode collapse, we additionally evaluate the free
energy $ F=-k_\text{B} T \ln p(\phi_i, \psi_i) $ of the backbone dihedral angles
(Ramachandran plots). Since these are the main slow degrees of freedom of the
peptide systems, mode collapse will be directly visible here. To assess the
quality of the Ramachandran plot, we calculate the forward KLD between the
probability distribution given by the Ramachandran plot of the ground truth and
the one of the flow distribution (RAM KLD). Since the tetrapeptide and
hexapeptide have multiple pairs of backbone dihedral angles, we report the mean
of their RAM KLD values. Furthermore, we repeat the calculation of the RAM KLD
also using the Ramachandran plots obtained from importance sampling. 

Last, to evaluate the sampling efficiency, we report the reverse effective
sample size (ESS) (see Section~\ref{sec:preliminaries}).

\section{Results}
Figure~\ref{main:fig_annealing} shows how the Ramachandran plots of each of the
three systems are annealed to the target temperature, showing four exemplary
steps of the annealing workflow. Both the distribution at \SI{1200}{\kelvin}
learned with the reverse KLD and the distribution at the target temperature
\SI{300}{\kelvin} match the ground truth obtained from MD.

\begin{figure*}[ht]
\vskip 0.2in
\begin{center}
\centerline{\includegraphics{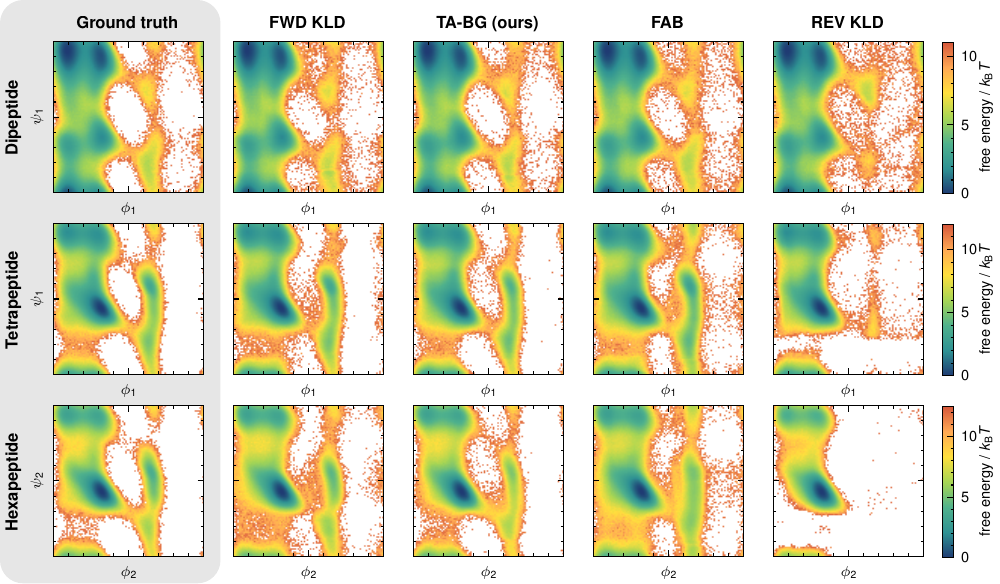}} \caption{Comparison of
the free energy $ F=-k_\text{B} T \ln p(\phi_i, \psi_i) $ of the backbone
dihedral angles (Ramachandran plots) at \SI{300}{\kelvin} obtained by the
different methods. Since the tetrapeptide has three pairs of backbone dihedral
angles, and the hexapeptide five, we selected the pair with the largest visible
deviation among the methods. The Ramachandran plots for all pairs of backbone
dihedral angles can be found in the appendix,
Figures~\ref{SI:tetra_all_ramachandrans} and \ref{SI:hexa_all_ramachandrans}.}
\label{main:fig_ram_comparison}
\end{center}
\vskip -0.2in
\end{figure*}

We now compare the obtained distribution at \SI{300}{\kelvin} with that from the
baseline methods by evaluating the introduced metrics
(Table~\ref{tab:main_results}). Furthermore,
Figure~\ref{main:fig_ram_comparison} shows the Ramachandran plots at
\SI{300}{\kelvin} obtained by all methods side-by-side.
Figure~\ref{SI:fig_ram_comparison_reweighted} in the appendix shows the same
comparison, but with importance sampling to the target distribution.

We start with the smallest system, alanine dipeptide. All methods, except for the reverse KLD, are able to learn
the distribution without mode collapse (Figure~\ref{main:fig_ram_comparison}). 
While the reverse KLD covers the high-energy region to some extent,
partial mode collapse is visible. In terms of metrics, FAB and TA-BG obtain comparable results. 
Our method achieves better NLL and ESS values, while FAB achieves slightly lower RAM KLD values.
Our approach only uses approximately a third of the target energy evaluations of FAB.
We further note that 
the metrics we obtained with FAB on alanine dipeptide are slightly better but mostly comparable to those in the original publication.

Similar results can be observed for the tetrapeptide system. The reverse KLD training now collapses almost fully
to the main mode, missing most of the metastable region (Figure~\ref{main:fig_ram_comparison}). 
Both TA-BG and FAB achieve a good match with the ground 
truth distribution, fully resolving the metastable region. The metrics of TA-BG and FAB are again close, our 
approach achieves slightly better NLL and RAM KLD values, while FAB has a lower reweighted RAM KLD value and slightly
higher ESS. Our approach again uses approximately a third of the target energy evaluations of FAB.

In the case of the most complex investigated system, alanine hexapeptide, the distribution of the reverse KLD training 
again almost entirely collapses. While FAB is partially able to resolve the metastable states,
they differ in shape compared to the ground truth distribution. In contrast, our approach covers all metastable states without mode collapse 
and resolves them accurately with only small imperfections. This is also reflected and quantified by the metrics: Compared to FAB, our approach achieves better results in all metrics, while requiring \num{3.08e8} target energy evaluations compared to \num{4.2e8} used by FAB.

A tradeoff exists between the accuracy of the obtained distribution and the number of target evaluations. This is especially true for FAB, where the number of intermediate AIS distributions and the number of HMC steps can be varied. We present corresponding variations in Table~\ref{SI:FAB_variations} in the appendix. For FAB applied to the hexapeptide, even when using almost 3 times as many target evaluations compared to our approach, we still achieve a lower NLL value.



\begin{table*}[t]
\caption{Comparison of metrics obtained for all three peptide systems. 
The table shows the number of potential energy evaluations (PE EVALS), the negative log-likelihood (NLL), the 
reverse effective sample size (ESS), the (mean) Ramachandran KLD (RAM KLD), and the (mean) 
Ramachandran KLD with reweighting (RAM KLD W. RW). All metrics are reported as the mean and standard error obtained from 
four independent experiments. The best-performing variational method for each metric is highlighted in bold.}
\label{tab:main_results}
\vskip 0.15in
\begin{center}
\begin{small}
\begin{sc}
\resizebox{\textwidth}{!}{\begin{tabular}{clccccc}
\toprule
System & Method & PE EVALS $\downarrow$ & NLL $\downarrow$ & ESS $\uparrow$ & Ram KLD $\downarrow$ & Ram KLD w. RW $\downarrow$ \\
\midrule

\multirow{5}{*}{dipeptide} 
                             & forward KLD & \num{5e9} & \num{-213.581(0.000)e0} & \SI{82.16(0.09)}{\percent} & \num{2.21(0.05)e-3} & \num{1.99(0.07)e-3} \\
\cmidrule(lr){2-7}
                             & reverse KLD & \num{2.56e8} & \num{-213.609(0.006)e0} & \SI{94.11(0.21)}{\percent} & \num{1.75(0.28)e-2} & \num{1.65(0.29)e-2} \\
                             & FAB & \num{2.13e8} & \num{-213.653(0.000)e0} & \SI{94.81(0.04)}{\percent} & \textbf{\num{1.50(0.03)e-3}} & \textbf{\num{1.25(0.01)e-3}} \\
                             & TA-BG (ours) & \textbf{\num{7.56e7}} & \textbf{\num{-213.665(0.002)e0}} & \textbf{\SI{95.61(0.14)}{\percent}} & \num{1.97(0.08)e-3} & \num{1.39(0.03)e-3} \\

\midrule

\multirow{5}{*}{\shortstack[c]{Tetra- \\ peptide}} 
                             & forward KLD & \num{4.2e9} & \num{-330.069(0.001)e0} & \SI{45.29(0.08)}{\percent} & \num{2.26(0.06)e-3} & \num{2.50(0.03)e-3} \\
\cmidrule(lr){2-7}
                             & reverse KLD & \num{2.56e8} & \num{-329.191(0.122)e0} & \textbf{\SI{74.88(3.65)}{\percent}} & \num{3.00(0.35)e-1} & \num{2.87(0.40)e-1} \\
                             & FAB & \num{2.13e8} & \num{-330.100(0.002)e0} & \SI{63.59(0.23)}{\percent} & \num{6.89(0.25)e-3} & \textbf{\num{1.25(0.01)e-3}} \\
                             & TA-BG (ours) & \textbf{\num{7.56e7}} & \textbf{\num{-330.114(0.004)e0}} & \SI{62.36(0.46)}{\percent} & \textbf{\num{2.61(0.20)e-3}} & \num{1.94(0.12)e-3} \\

\midrule

\multirow{5}{*}{\shortstack[c]{Hexa- \\ peptide}} 
                             & forward KLD & \num{4.2e9} & \num{-501.598(0.005)e0} & \SI{10.97(0.11)}{\percent} & \num{4.16(0.26)e-3} & \num{7.69(0.03)e-3} \\
\cmidrule(lr){2-7}
                             & reverse KLD & \textbf{\num{2.56e8}} & \num{-497.378(0.277)e0} & \textbf{\SI{22.22(1.44)}{\percent}} & \num{5.41(0.38)e-1} & \num{5.32(0.38)e-1} \\
                             & FAB & \num{4.2e8} & \num{-501.268(0.008)e0} & \SI{14.64(0.08)}{\percent} & \num{2.09(0.02)e-2} & \num{1.12(0.02)e-2} \\
                             & TA-BG (ours) & \num{3.08e8} & \textbf{\num{-501.511(0.013)e0}} & \SI{14.71(0.18)}{\percent} & \textbf{\num{8.63(0.66)e-3}} & \textbf{\num{8.77(0.47)e-3}} \\

\bottomrule
\end{tabular}}
\end{sc}
\end{small}
\label{main:tab1}
\end{center}
\vskip -0.1in
\end{table*}

\section{Discussion}
To summarize, TA-BG achieves better results in most metrics compared
to the baselines and thus learns the Boltzmann distribution of the investigated
systems more accurately, while requiring significantly fewer target energy
evaluations. We note that force field evaluations in our current setup are
relatively inexpensive. With more accurate and computationally costly target
evaluations, such as ML-based foundation models or density functional theory,
the energy evaluation cost becomes dominant. In such cases, the higher sampling
efficiency of our method will translate into significantly reduced computational
cost. We refer to Section~\ref{SI:sec_comp_cost} in the appendix for a detailed
analysis of the computational cost of TA-BG in comparison with FAB.

Next, we would like to point out the surprisingly good results we obtained by
simply training with the reverse KLD at \SI{300}{\kelvin} for alanine dipeptide
(Figure~\ref{main:fig_ram_comparison}). Even though at \SI{300}{\kelvin} the
metastable states are only connected to the global minimum by a very
low-probability transition region, only partial mode collapse is observed. Next
to our results of being able to successfully learn the distribution at high
temperatures with the reverse KLD, this further shows the effectiveness of
simple reverse KLD training.

Furthermore, previous work showed that not only training a normalizing flow from
scratch with the reverse KLD is prone to mode collapse, but also fine-tuning a
pre-trained flow to a target distribution with the reverse KLD typically
collapses \cite{felardosDesigningLossesDatafree2023}. This is a
not well understood phenomenon and has only recently been investigated
theoretically \cite{soletskyiTheoreticalPerspectiveMode2024}. While
\citeauthor{felardosDesigningLossesDatafree2023}
\yrcite{felardosDesigningLossesDatafree2023} proposed a solution to this
problem, it is not clear how well it scales to larger systems. As described
before, after our annealing workflow reaches \SI{300}{\kelvin}, we fine-tune the
flow distribution at \SI{300}{\kelvin} by performing forward KLD training with a
buffered dataset from the flow distribution, reweighted to the target
distribution. This offers a simple yet effective solution to the problem of
fine-tuning pre-trained flows and can also be used in other scenarios. For
example, one can train a flow on biased MD simulation data that is not properly
equilibrated, and then fine-tune with our buffered reweighting approach to
obtain an unbiased flow distribution.

Furthermore, while our method generally yielded better results with fewer target
energy evaluations for the hexapeptide, the results we obtained using FAB are
still relatively close in terms of accuracy, especially for the other two
systems. This establishes it as a powerful method for variational sampling.
Therefore, a combination of our temperature-annealing approach with FAB is an
interesting avenue for future work. Instead of using the reverse KLD for
training at increased temperatures, FAB can be used and the distribution can be
subsequently annealed with our TA-BG workflow. If and when this can be better
than simple reverse KLD training at high temperatures needs to be investigated.

Our results indicate that other variational sampling approaches, such as
those based on diffusion models, might benefit from sampling at higher
temperatures and annealing the temperature afterwards. Other objectives will
likely also benefit from the reduced mode separation at elevated temperatures.

In this work, we used a relatively simple flow architecture based on neural
splines and internal coordinates. Recent improvements in flow architectures
\cite{zhaiNormalizingFlowsAre2024,tanScalableEquilibriumSampling2025} can be
leveraged in the future to improve the obtained results and scale to larger
systems. Furthermore, the use of an internal coordinate representation can be a
limitation, since it is not transferable between systems and it does not incorporate 
the permutation invariance of identical particles. A transferable
equivariant normalizing flow has recently been proposed
\cite{midgleySE3EquivariantAugmented2023}. Using the TA-BG methodology in a
transferable setting is an interesting avenue for future work, which may
also allow a hybrid approach where data-free training is combined with
data-based training \cite{lewisScalableEmulationProtein2024}.

\section{Conclusion}
We introduced temperature-annealed Boltzmann generators, a technique that uses a
combination of high-temperature reverse KLD pre-training and a subsequent annealing workflow
to efficiently sample the Boltzmann distribution of molecular systems at room
temperature. On the molecular systems investigated, our approach achieves better
results in almost all metrics, while requiring up to three times fewer target
energy evaluations compared to the baselines. Furthermore, it was the only
variational approach that accurately resolved the metastable region of the most
complex system studied, demonstrating its scaling capabilities from toy examples
to application-relevant systems.

Similar to how replica exchange molecular dynamics is an established method in
the toolbox of computational scientists, we are confident that high-temperature
sampling with temperature-annealing is a powerful approach that will move the
field of variational sampling forward.

\section*{Software and Data}
The source code to reproduce our experiments can be found at \url{https://github.com/aimat-lab/TA-BG}.
Furthermore, the ground truth datasets from MD simulations are 
provided at \url{https://doi.org/10.5281/zenodo.15526429}.

\section*{Acknowledgments}
The authors would like to thank the anonymous reviewers for their valuable
comments and suggestions. H.S. acknowledges financial support by the German
Research Foundation (DFG) through the Research Training Group 2450 “Tailored
Scale-Bridging Approaches to Computational Nanoscience”.
P.F. acknowledges funding from the Klaus Tschira Stiftung gGmbH (SIMPLAIX project) and the pilot program Core-Informatics of the Helmholtz Association (KiKIT project).
Parts of this work were
performed on the HoreKa supercomputer funded by the Ministry of Science,
Research and the Arts Baden-Württemberg and by the Federal Ministry of Education
and Research. This work is supported by the Helmholtz Association Initiative and
Networking Fund on the HAICORE@KIT partition.

\section*{Impact Statement}
This paper presents work whose goal is to advance the field of 
Machine Learning. There are many potential societal consequences 
of our work, none which we feel must be specifically highlighted here.

\clearpage

\bibliography{main}

\begin{thebibliography}{52}
\providecommand{\natexlab}[1]{#1}
\providecommand{\url}[1]{\texttt{#1}}
\expandafter\ifx\csname urlstyle\endcsname\relax
  \providecommand{\doi}[1]{doi: #1}\else
  \providecommand{\doi}{doi: \begingroup \urlstyle{rm}\Url}\fi

\bibitem[Abramson et~al.(2024)Abramson, Adler, Dunger, Evans, Green, Pritzel,
  Ronneberger, Willmore, Ballard, Bambrick, Bodenstein, Evans, Hung, O'Neill,
  Reiman, Tunyasuvunakool, Wu, {\v Z}emgulyt{\.e}, Arvaniti, Beattie, Bertolli,
  Bridgland, Cherepanov, Congreve, {Cowen-Rivers}, Cowie, Figurnov, Fuchs,
  Gladman, Jain, Khan, Low, Perlin, Potapenko, Savy, Singh, Stecula,
  Thillaisundaram, Tong, Yakneen, Zhong, Zielinski, {\v Z}{\'i}dek, Bapst,
  Kohli, Jaderberg, Hassabis, and
  Jumper]{abramsonAccurateStructurePrediction2024}
Abramson, J., Adler, J., Dunger, J., Evans, R., Green, T., Pritzel, A.,
  Ronneberger, O., Willmore, L., Ballard, A.~J., Bambrick, J., Bodenstein,
  S.~W., Evans, D.~A., Hung, C.-C., O'Neill, M., Reiman, D., Tunyasuvunakool,
  K., Wu, Z., {\v Z}emgulyt{\.e}, A., Arvaniti, E., Beattie, C., Bertolli, O.,
  Bridgland, A., Cherepanov, A., Congreve, M., {Cowen-Rivers}, A.~I., Cowie,
  A., Figurnov, M., Fuchs, F.~B., Gladman, H., Jain, R., Khan, Y.~A., Low, C.
  M.~R., Perlin, K., Potapenko, A., Savy, P., Singh, S., Stecula, A.,
  Thillaisundaram, A., Tong, C., Yakneen, S., Zhong, E.~D., Zielinski, M., {\v
  Z}{\'i}dek, A., Bapst, V., Kohli, P., Jaderberg, M., Hassabis, D., and
  Jumper, J.~M.
\newblock Accurate structure prediction of biomolecular interactions with
  {{AlphaFold}} 3.
\newblock \emph{Nature}, 630\penalty0 (8016):\penalty0 493--500, June 2024.
\newblock ISSN 1476-4687.
\newblock \doi{10.1038/s41586-024-07487-w}.

\bibitem[{Akhound-Sadegh} et~al.(2024){Akhound-Sadegh}, {Rector-Brooks}, Bose,
  Mittal, Lemos, Liu, Sendera, Ravanbakhsh, Gidel, Bengio, Malkin, and
  Tong]{akhound-sadeghIteratedDenoisingEnergy2024a}
{Akhound-Sadegh}, T., {Rector-Brooks}, J., Bose, J., Mittal, S., Lemos, P.,
  Liu, C.-H., Sendera, M., Ravanbakhsh, S., Gidel, G., Bengio, Y., Malkin, N.,
  and Tong, A.
\newblock Iterated {{Denoising Energy Matching}} for {{Sampling}} from
  {{Boltzmann Densities}}.
\newblock In \emph{Forty-First {{International Conference}} on {{Machine
  Learning}}}, June 2024.

\bibitem[Barducci et~al.(2011)Barducci, Bonomi, and
  Parrinello]{barducciMetadynamics2011}
Barducci, A., Bonomi, M., and Parrinello, M.
\newblock Metadynamics.
\newblock \emph{WIREs Computational Molecular Science}, 1\penalty0
  (5):\penalty0 826--843, 2011.
\newblock ISSN 1759-0884.
\newblock \doi{10.1002/wcms.31}.

\bibitem[Berner et~al.(2023)Berner, Richter, and
  Ullrich]{bernerOptimalControlPerspective2023}
Berner, J., Richter, L., and Ullrich, K.
\newblock An optimal control perspective on diffusion-based generative
  modeling.
\newblock \emph{Transactions on Machine Learning Research}, October 2023.
\newblock ISSN 2835-8856.

\bibitem[Blessing et~al.(2024)Blessing, Jia, Esslinger, Vargas, and
  Neumann]{blessingELBOsLargeScaleEvaluation2024}
Blessing, D., Jia, X., Esslinger, J., Vargas, F., and Neumann, G.
\newblock Beyond {{ELBOs}}: {{A Large-Scale Evaluation}} of {{Variational
  Methods}} for {{Sampling}}.
\newblock In \emph{Forty-First {{International Conference}} on {{Machine
  Learning}}}, June 2024.

\bibitem[Bonati et~al.(2021)Bonati, Piccini, and
  Parrinello]{bonatiDeepLearningSlow2021}
Bonati, L., Piccini, G., and Parrinello, M.
\newblock Deep learning the slow modes for rare events sampling.
\newblock \emph{Proceedings of the National Academy of Sciences}, 118\penalty0
  (44):\penalty0 e2113533118, November 2021.
\newblock \doi{10.1073/pnas.2113533118}.

\bibitem[Burley et~al.(2021)Burley, Bhikadiya, Bi, Bittrich, Chen, Crichlow,
  Christie, Dalenberg, Di~Costanzo, Duarte, Dutta, Feng, Ganesan, Goodsell,
  Ghosh, Green, Guranovi{\'c}, Guzenko, Hudson, Lawson, Liang, Lowe, Namkoong,
  Peisach, Persikova, Randle, Rose, Rose, Sali, Segura, Sekharan, Shao, Tao,
  Voigt, Westbrook, Young, Zardecki, and Zhuravleva]{burleyRCSBProteinData2021}
Burley, S.~K., Bhikadiya, C., Bi, C., Bittrich, S., Chen, L., Crichlow, G.~V.,
  Christie, C.~H., Dalenberg, K., Di~Costanzo, L., Duarte, J.~M., Dutta, S.,
  Feng, Z., Ganesan, S., Goodsell, D.~S., Ghosh, S., Green, R.~K.,
  Guranovi{\'c}, V., Guzenko, D., Hudson, B.~P., Lawson, C.~L., Liang, Y.,
  Lowe, R., Namkoong, H., Peisach, E., Persikova, I., Randle, C., Rose, A.,
  Rose, Y., Sali, A., Segura, J., Sekharan, M., Shao, C., Tao, Y.-P., Voigt,
  M., Westbrook, J.~D., Young, J.~Y., Zardecki, C., and Zhuravleva, M.
\newblock {{RCSB Protein Data Bank}}: Powerful new tools for exploring {{3D}}
  structures of biological macromolecules for basic and applied research and
  education in fundamental biology, biomedicine, biotechnology, bioengineering
  and energy sciences.
\newblock \emph{Nucleic Acids Research}, 49\penalty0 (D1):\penalty0 D437--D451,
  January 2021.
\newblock ISSN 0305-1048.
\newblock \doi{10.1093/nar/gkaa1038}.

\bibitem[{Conor Durkan} et~al.(2020){Conor Durkan}, {Artur Bekasov}, {Iain
  Murray}, and {George Papamakarios}]{conordurkanNflowsNormalizingFlows2020}
{Conor Durkan}, {Artur Bekasov}, {Iain Murray}, and {George Papamakarios}.
\newblock Nflows: Normalizing flows in {{PyTorch}}, November 2020.

\bibitem[{D.A. Case} et~al.(2023){D.A. Case}, {H.M. Aktulga}, {K. Belfon},
  {I.Y. Ben-Shalom}, {J.T. Berryman}, {S.R. Brozell}, {D.S. Cerutti}, {T.E.
  Cheatham, III}, {V.W.D. Cruzeiro}, {T.A. Darden}, {N. Forouzesh}, {G.
  Giambasu}, {T. Giese}, {M.K. Gilson}, {H. Gohlke}, {A.W. Goetz}, {J. Harris},
  {S. Izadi}, {S.A. Izmailov}, {K. Kasavajhala}, {M.C. Kaymak}, {E. King}, {A.
  Kovalenko}, {T. Kurtzman}, {T.S. Lee}, {P. Li}, {C. Lin}, {J. Liu}, {T.
  Luchko}, {R. Luo}, {M. Machado}, {V. Man}, {M. Manathunga}, {K.M. Merz}, {Y.
  Miao}, {O. Mikhailovskii}, {G. Monard}, {H. Nguyen}, {K.A. O'Hearn}, {A.
  Onufriev}, {F. Pan}, {S. Pantano}, {R. Qi}, {A. Rahnamoun}, {D.R. Roe}, {A.
  Roitberg}, {C. Sagui}, {S. Schott-Verdugo}, {A. Shajan}, {J. Shen}, {C.L.
  Simmerling}, {N.R. Skrynnikov}, {J. Smith}, {J. Swails}, {R.C. Walker}, {J.
  Wang}, {J. Wang}, {H. Wei}, {X. Wu}, {Y. Wu}, {Y. Xiong}, {Y. Xue}, {D.M.
  York}, {S. Zhao}, {Q. Zhu}, and {P.A. Kollman}]{d.a.caseAmber20232023}
{D.A. Case}, {H.M. Aktulga}, {K. Belfon}, {I.Y. Ben-Shalom}, {J.T. Berryman},
  {S.R. Brozell}, {D.S. Cerutti}, {T.E. Cheatham, III}, {V.W.D. Cruzeiro},
  {T.A. Darden}, {N. Forouzesh}, {G. Giambasu}, {T. Giese}, {M.K. Gilson}, {H.
  Gohlke}, {A.W. Goetz}, {J. Harris}, {S. Izadi}, {S.A. Izmailov}, {K.
  Kasavajhala}, {M.C. Kaymak}, {E. King}, {A. Kovalenko}, {T. Kurtzman}, {T.S.
  Lee}, {P. Li}, {C. Lin}, {J. Liu}, {T. Luchko}, {R. Luo}, {M. Machado}, {V.
  Man}, {M. Manathunga}, {K.M. Merz}, {Y. Miao}, {O. Mikhailovskii}, {G.
  Monard}, {H. Nguyen}, {K.A. O'Hearn}, {A. Onufriev}, {F. Pan}, {S. Pantano},
  {R. Qi}, {A. Rahnamoun}, {D.R. Roe}, {A. Roitberg}, {C. Sagui}, {S.
  Schott-Verdugo}, {A. Shajan}, {J. Shen}, {C.L. Simmerling}, {N.R.
  Skrynnikov}, {J. Smith}, {J. Swails}, {R.C. Walker}, {J. Wang}, {J. Wang},
  {H. Wei}, {X. Wu}, {Y. Wu}, {Y. Xiong}, {Y. Xue}, {D.M. York}, {S. Zhao}, {Q.
  Zhu}, and {P.A. Kollman}.
\newblock Amber 2023.
\newblock University of California, San Francisco, 2023.

\bibitem[Dibak et~al.(2022)Dibak, Klein, Kr{\"a}mer, and
  No{\'e}]{dibakTemperatureSteerableFlows2022}
Dibak, M., Klein, L., Kr{\"a}mer, A., and No{\'e}, F.
\newblock Temperature steerable flows and {{Boltzmann}} generators.
\newblock \emph{Phys. Rev. Res.}, 4\penalty0 (4):\penalty0 L042005, October
  2022.
\newblock \doi{10.1103/PhysRevResearch.4.L042005}.

\bibitem[Dinh et~al.(2015)Dinh, Krueger, and
  Bengio]{dinhNICENonlinearIndependent2015}
Dinh, L., Krueger, D., and Bengio, Y.
\newblock {{NICE}}: {{Non-linear Independent Components Estimation}}, April
  2015.

\bibitem[Dinh et~al.(2017)Dinh, {Sohl-Dickstein}, and
  Bengio]{dinhDensityEstimationUsing2017}
Dinh, L., {Sohl-Dickstein}, J., and Bengio, S.
\newblock Density estimation using {{Real NVP}}, February 2017.

\bibitem[Draxler et~al.(2024)Draxler, Wahl, Schnoerr, and
  Koethe]{draxlerUniversalityVolumePreservingCouplingBased2024}
Draxler, F., Wahl, S., Schnoerr, C., and Koethe, U.
\newblock On the {{Universality}} of {{Volume-Preserving}} and {{Coupling-Based
  Normalizing Flows}}.
\newblock In \emph{Forty-First {{International Conference}} on {{Machine
  Learning}}}, June 2024.

\bibitem[Duane et~al.(1987)Duane, Kennedy, Pendleton, and
  Roweth]{duaneHybridMonteCarlo1987}
Duane, S., Kennedy, A.~D., Pendleton, B.~J., and Roweth, D.
\newblock Hybrid {{Monte Carlo}}.
\newblock \emph{Physics Letters B}, 195\penalty0 (2):\penalty0 216--222,
  September 1987.
\newblock ISSN 0370-2693.
\newblock \doi{10.1016/0370-2693(87)91197-X}.

\bibitem[Durkan et~al.(2019)Durkan, Bekasov, Murray, and
  Papamakarios]{durkanNeuralSplineFlows2019a}
Durkan, C., Bekasov, A., Murray, I., and Papamakarios, G.
\newblock Neural {{Spline Flows}}.
\newblock In \emph{Advances in {{Neural Information Processing Systems}}},
  volume~32. Curran Associates, Inc., 2019.

\bibitem[Eastman et~al.(2024)Eastman, Galvelis, Pel{\'a}ez, Abreu, Farr,
  Gallicchio, Gorenko, Henry, Hu, Huang, Kr{\"a}mer, Michel, Mitchell, Pande,
  Rodrigues, {Rodriguez-Guerra}, Simmonett, Singh, Swails, Turner, Wang, Zhang,
  Chodera, De~Fabritiis, and Markland]{eastmanOpenMM8Molecular2024}
Eastman, P., Galvelis, R., Pel{\'a}ez, R.~P., Abreu, C. R.~A., Farr, S.~E.,
  Gallicchio, E., Gorenko, A., Henry, M.~M., Hu, F., Huang, J., Kr{\"a}mer, A.,
  Michel, J., Mitchell, J.~A., Pande, V.~S., Rodrigues, J.~P.,
  {Rodriguez-Guerra}, J., Simmonett, A.~C., Singh, S., Swails, J., Turner, P.,
  Wang, Y., Zhang, I., Chodera, J.~D., De~Fabritiis, G., and Markland, T.~E.
\newblock {{OpenMM}} 8: {{Molecular Dynamics Simulation}} with {{Machine
  Learning Potentials}}.
\newblock \emph{J. Phys. Chem. B}, 128\penalty0 (1):\penalty0 109--116, January
  2024.
\newblock ISSN 1520-6106.
\newblock \doi{10.1021/acs.jpcb.3c06662}.

\bibitem[Felardos et~al.(2023)Felardos, H{\'e}nin, and
  Charpiat]{felardosDesigningLossesDatafree2023}
Felardos, L., H{\'e}nin, J., and Charpiat, G.
\newblock Designing losses for data-free training of normalizing flows on
  {{Boltzmann}} distributions, January 2023.

\bibitem[Goshtasbpour et~al.(2023)Goshtasbpour, Cohen, and
  {Perez-Cruz}]{goshtasbpourAdaptiveAnnealedImportance2023}
Goshtasbpour, S., Cohen, V., and {Perez-Cruz}, F.
\newblock Adaptive {{Annealed Importance Sampling}} with {{Constant Rate
  Progress}}.
\newblock In \emph{Proceedings of the 40th {{International Conference}} on
  {{Machine Learning}}}, pp.\  11642--11658. PMLR, July 2023.

\bibitem[Holdijk et~al.(2023)Holdijk, Du, Hooft, Jaini, Ensing, and
  Welling]{holdijkStochasticOptimalControl2023}
Holdijk, L., Du, Y., Hooft, F., Jaini, P., Ensing, B., and Welling, M.
\newblock Stochastic {{Optimal Control}} for {{Collective Variable Free
  Sampling}} of {{Molecular Transition Paths}}.
\newblock \emph{Advances in Neural Information Processing Systems},
  36:\penalty0 79540--79556, December 2023.

\bibitem[Invernizzi et~al.(2022)Invernizzi, Kr{\"a}mer, Clementi, and
  No{\'e}]{invernizziSkippingReplicaExchange2022}
Invernizzi, M., Kr{\"a}mer, A., Clementi, C., and No{\'e}, F.
\newblock Skipping the {{Replica Exchange Ladder}} with {{Normalizing Flows}}.
\newblock \emph{J. Phys. Chem. Lett.}, 13\penalty0 (50):\penalty0 11643--11649,
  December 2022.
\newblock \doi{10.1021/acs.jpclett.2c03327}.

\bibitem[Jumper et~al.(2021)Jumper, Evans, Pritzel, Green, Figurnov,
  Ronneberger, Tunyasuvunakool, Bates, {\v Z}{\'i}dek, Potapenko, Bridgland,
  Meyer, Kohl, Ballard, Cowie, {Romera-Paredes}, Nikolov, Jain, Adler, Back,
  Petersen, Reiman, Clancy, Zielinski, Steinegger, Pacholska, Berghammer,
  Bodenstein, Silver, Vinyals, Senior, Kavukcuoglu, Kohli, and
  Hassabis]{jumperHighlyAccurateProtein2021}
Jumper, J., Evans, R., Pritzel, A., Green, T., Figurnov, M., Ronneberger, O.,
  Tunyasuvunakool, K., Bates, R., {\v Z}{\'i}dek, A., Potapenko, A., Bridgland,
  A., Meyer, C., Kohl, S. A.~A., Ballard, A.~J., Cowie, A., {Romera-Paredes},
  B., Nikolov, S., Jain, R., Adler, J., Back, T., Petersen, S., Reiman, D.,
  Clancy, E., Zielinski, M., Steinegger, M., Pacholska, M., Berghammer, T.,
  Bodenstein, S., Silver, D., Vinyals, O., Senior, A.~W., Kavukcuoglu, K.,
  Kohli, P., and Hassabis, D.
\newblock Highly accurate protein structure prediction with {{AlphaFold}}.
\newblock \emph{Nature}, 596\penalty0 (7873):\penalty0 583--589, August 2021.
\newblock ISSN 1476-4687.
\newblock \doi{10.1038/s41586-021-03819-2}.

\bibitem[Kingma \& Ba(2017)Kingma and Ba]{kingmaAdamMethodStochastic2017}
Kingma, D.~P. and Ba, J.
\newblock Adam: {{A Method}} for {{Stochastic Optimization}}, January 2017.

\bibitem[Klein \& Noe(2024)Klein and
  Noe]{kleinTransferableBoltzmannGenerators2024a}
Klein, L. and Noe, F.
\newblock Transferable {{Boltzmann Generators}}.
\newblock In \emph{The {{Thirty-eighth Annual Conference}} on {{Neural
  Information Processing Systems}}}, November 2024.

\bibitem[Lewis et~al.(2024)Lewis, Hempel, {Jim{\'e}nez-Luna}, Gastegger, Xie,
  Foong, Satorras, Abdin, Veeling, Zaporozhets, Chen, Yang, Schneuing, Nigam,
  Barbero, Stimper, Campbell, Yim, Lienen, Shi, Zheng, Schulz, Munir, Clementi,
  and No{\'e}]{lewisScalableEmulationProtein2024}
Lewis, S., Hempel, T., {Jim{\'e}nez-Luna}, J., Gastegger, M., Xie, Y., Foong,
  A. Y.~K., Satorras, V.~G., Abdin, O., Veeling, B.~S., Zaporozhets, I., Chen,
  Y., Yang, S., Schneuing, A., Nigam, J., Barbero, F., Stimper, V., Campbell,
  A., Yim, J., Lienen, M., Shi, Y., Zheng, S., Schulz, H., Munir, U., Clementi,
  C., and No{\'e}, F.
\newblock Scalable emulation of protein equilibrium ensembles with generative
  deep learning, December 2024.

\bibitem[Mahmoud et~al.(2022)Mahmoud, Masters, Lee, and
  Lill]{mahmoudAccurateSamplingMacromolecular2022}
Mahmoud, A.~H., Masters, M., Lee, S.~J., and Lill, M.~A.
\newblock Accurate {{Sampling}} of {{Macromolecular Conformations Using
  Adaptive Deep Learning}} and {{Coarse-Grained Representation}}.
\newblock \emph{J. Chem. Inf. Model.}, 62\penalty0 (7):\penalty0 1602--1617,
  April 2022.
\newblock ISSN 1549-9596.
\newblock \doi{10.1021/acs.jcim.1c01438}.

\bibitem[Martino et~al.(2017)Martino, Elvira, and
  Louzada]{martinoEffectiveSampleSize2017}
Martino, L., Elvira, V., and Louzada, F.
\newblock Effective {{Sample Size}} for {{Importance Sampling}} based on
  discrepancy measures.
\newblock \emph{Signal Processing}, 131:\penalty0 386--401, February 2017.
\newblock ISSN 01651684.
\newblock \doi{10.1016/j.sigpro.2016.08.025}.

\bibitem[Matthews et~al.(2022)Matthews, Arbel, Rezende, and
  Doucet]{matthewsContinualRepeatedAnnealed2022}
Matthews, A., Arbel, M., Rezende, D.~J., and Doucet, A.
\newblock Continual {{Repeated Annealed Flow Transport Monte Carlo}}.
\newblock In \emph{Proceedings of the 39th {{International Conference}} on
  {{Machine Learning}}}, pp.\  15196--15219. PMLR, June 2022.

\bibitem[Midgley et~al.(2023{\natexlab{a}})Midgley, Stimper, Antor{\'a}n,
  Mathieu, Sch{\"o}lkopf, and
  {Hern{\'a}ndez-Lobato}]{midgleySE3EquivariantAugmented2023}
Midgley, L.~I., Stimper, V., Antor{\'a}n, J., Mathieu, E., Sch{\"o}lkopf, B.,
  and {Hern{\'a}ndez-Lobato}, J.~M.
\newblock {{SE}}(3) {{Equivariant Augmented Coupling Flows}}.
\newblock In \emph{Thirty-Seventh {{Conference}} on {{Neural Information
  Processing Systems}}}, August 2023{\natexlab{a}}.

\bibitem[Midgley et~al.(2023{\natexlab{b}})Midgley, Stimper, Simm,
  Sch{\"o}lkopf, and {Hern{\'a}ndez-Lobato}]{midgleyFlowAnnealedImportance2023}
Midgley, L.~I., Stimper, V., Simm, G. N.~C., Sch{\"o}lkopf, B., and
  {Hern{\'a}ndez-Lobato}, J.~M.
\newblock Flow {{Annealed Importance Sampling Bootstrap}}.
\newblock In \emph{The {{Eleventh International Conference}} on {{Learning
  Representations}}}, 2023{\natexlab{b}}.
\newblock \doi{10.48550/arXiv.2208.01893}.

\bibitem[Neal(2001)]{nealAnnealedImportanceSampling2001}
Neal, R.~M.
\newblock Annealed importance sampling.
\newblock \emph{Statistics and Computing}, 11\penalty0 (2):\penalty0 125--139,
  April 2001.
\newblock ISSN 1573-1375.
\newblock \doi{10.1023/A:1008923215028}.

\bibitem[No{\'e}(2024)]{noeBgflow2024}
No{\'e}, F.
\newblock Bgflow.
\newblock AI4Science group, FU Berlin (Frank No{\'e} and co-workers), December
  2024.

\bibitem[No{\'e} et~al.(2019)No{\'e}, Olsson, K{\"o}hler, and
  Wu]{noeBoltzmannGeneratorsSampling2019a}
No{\'e}, F., Olsson, S., K{\"o}hler, J., and Wu, H.
\newblock Boltzmann generators: {{Sampling}} equilibrium states of many-body
  systems with deep learning.
\newblock \emph{Science}, 365\penalty0 (6457):\penalty0 eaaw1147, September
  2019.
\newblock \doi{10.1126/science.aaw1147}.

\bibitem[Paszke et~al.(2019)Paszke, Gross, Massa, Lerer, Bradbury, Chanan,
  Killeen, Lin, Gimelshein, Antiga, Desmaison, K{\"o}pf, Yang, DeVito, Raison,
  Tejani, Chilamkurthy, Steiner, Fang, Bai, and
  Chintala]{paszkePyTorchImperativeStyle2019}
Paszke, A., Gross, S., Massa, F., Lerer, A., Bradbury, J., Chanan, G., Killeen,
  T., Lin, Z., Gimelshein, N., Antiga, L., Desmaison, A., K{\"o}pf, A., Yang,
  E., DeVito, Z., Raison, M., Tejani, A., Chilamkurthy, S., Steiner, B., Fang,
  L., Bai, J., and Chintala, S.
\newblock {{PyTorch}}: {{An Imperative Style}}, {{High-Performance Deep
  Learning Library}}, December 2019.

\bibitem[Reiser et~al.(2022)Reiser, Neubert, Eberhard, Torresi, Zhou, Shao,
  Metni, {van Hoesel}, Schopmans, Sommer, and
  Friederich]{reiserGraphNeuralNetworks2022a}
Reiser, P., Neubert, M., Eberhard, A., Torresi, L., Zhou, C., Shao, C., Metni,
  H., {van Hoesel}, C., Schopmans, H., Sommer, T., and Friederich, P.
\newblock Graph neural networks for materials science and chemistry.
\newblock \emph{Commun Mater}, 3\penalty0 (1):\penalty0 1--18, November 2022.
\newblock ISSN 2662-4443.
\newblock \doi{10.1038/s43246-022-00315-6}.

\bibitem[Rezende et~al.(2020)Rezende, Papamakarios, Racaniere, Albergo, Kanwar,
  Shanahan, and Cranmer]{rezendeNormalizingFlowsTori2020a}
Rezende, D.~J., Papamakarios, G., Racaniere, S., Albergo, M., Kanwar, G.,
  Shanahan, P., and Cranmer, K.
\newblock Normalizing {{Flows}} on {{Tori}} and {{Spheres}}.
\newblock In \emph{Proceedings of the 37th {{International Conference}} on
  {{Machine Learning}}}, pp.\  8083--8092. PMLR, November 2020.

\bibitem[Richter \& Berner(2023)Richter and
  Berner]{richterImprovedSamplingLearned2023}
Richter, L. and Berner, J.
\newblock Improved sampling via learned diffusions.
\newblock In \emph{The {{Twelfth International Conference}} on {{Learning
  Representations}}}, October 2023.

\bibitem[Schebek et~al.(2024)Schebek, Invernizzi, No{\'e}, and
  Rogal]{schebekEfficientMappingPhase2024a}
Schebek, M., Invernizzi, M., No{\'e}, F., and Rogal, J.
\newblock Efficient mapping of phase diagrams with conditional {{Boltzmann
  Generators}}.
\newblock \emph{Mach. Learn.: Sci. Technol.}, 5\penalty0 (4):\penalty0 045045,
  November 2024.
\newblock ISSN 2632-2153.
\newblock \doi{10.1088/2632-2153/ad849d}.

\bibitem[Schopmans \& Friederich(2024)Schopmans and
  Friederich]{schopmansConditionalNormalizingFlows2024}
Schopmans, H. and Friederich, P.
\newblock Conditional {{Normalizing Flows}} for {{Active Learning}} of
  {{Coarse-Grained Molecular Representations}}.
\newblock In \emph{Forty-First {{International Conference}} on {{Machine
  Learning}}}, June 2024.

\bibitem[Sendera et~al.(2024)Sendera, Kim, Mittal, Lemos, Scimeca,
  {Rector-Brooks}, Adam, Bengio, and
  Malkin]{senderaImprovedOffpolicyTraining2024}
Sendera, M., Kim, M., Mittal, S., Lemos, P., Scimeca, L., {Rector-Brooks}, J.,
  Adam, A., Bengio, Y., and Malkin, N.
\newblock Improved off-policy training of diffusion samplers.
\newblock In \emph{The {{Thirty-eighth Annual Conference}} on {{Neural
  Information Processing Systems}}}, November 2024.

\bibitem[Seong et~al.(2024)Seong, Park, Kim, Kim, and
  Ahn]{seongTransitionPathSampling2024}
Seong, K., Park, S., Kim, S., Kim, W.~Y., and Ahn, S.
\newblock Transition {{Path Sampling}} with {{Improved Off-Policy Training}} of
  {{Diffusion Path Samplers}}.
\newblock In \emph{The {{Thirteenth International Conference}} on {{Learning
  Representations}}}, October 2024.

\bibitem[Soletskyi et~al.(2024)Soletskyi, Gabri{\'e}, and
  Loureiro]{soletskyiTheoreticalPerspectiveMode2024}
Soletskyi, R., Gabri{\'e}, M., and Loureiro, B.
\newblock A theoretical perspective on mode collapse in variational inference,
  October 2024.

\bibitem[Stimper et~al.(2022)Stimper, Midgley, Simm, Sch{\"o}lkopf, and
  {Hern{\'a}ndez-Lobato}]{stimperAlanineDipeptideImplicit2022}
Stimper, V., Midgley, L.~I., Simm, G. N.~C., Sch{\"o}lkopf, B., and
  {Hern{\'a}ndez-Lobato}, J.~M.
\newblock Alanine dipeptide in an implicit solvent at {{300K}}.
\newblock August 2022.
\newblock \doi{10.5281/zenodo.6993124}.

\bibitem[Sugita \& Okamoto(1999)Sugita and
  Okamoto]{sugitaReplicaexchangeMolecularDynamics1999}
Sugita, Y. and Okamoto, Y.
\newblock Replica-exchange molecular dynamics method for protein folding.
\newblock \emph{Chemical Physics Letters}, 314\penalty0 (1):\penalty0 141--151,
  November 1999.
\newblock ISSN 0009-2614.
\newblock \doi{10.1016/S0009-2614(99)01123-9}.

\bibitem[Tan et~al.(2025)Tan, Bose, Lin, Klein, Bronstein, and
  Tong]{tanScalableEquilibriumSampling2025}
Tan, C.~B., Bose, A.~J., Lin, C., Klein, L., Bronstein, M.~M., and Tong, A.
\newblock Scalable {{Equilibrium Sampling}} with {{Sequential Boltzmann
  Generators}}, February 2025.

\bibitem[Vargas et~al.(2022)Vargas, Grathwohl, and
  Doucet]{vargasDenoisingDiffusionSamplers2022}
Vargas, F., Grathwohl, W.~S., and Doucet, A.
\newblock Denoising {{Diffusion Samplers}}.
\newblock In \emph{The {{Eleventh International Conference}} on {{Learning
  Representations}}}, September 2022.

\bibitem[Vargas et~al.(2023)Vargas, Padhy, Blessing, and
  N{\"u}sken]{vargasTransportMeetsVariational2023}
Vargas, F., Padhy, S., Blessing, D., and N{\"u}sken, N.
\newblock Transport meets {{Variational Inference}}: {{Controlled Monte Carlo
  Diffusions}}.
\newblock In \emph{The {{Twelfth International Conference}} on {{Learning
  Representations}}}, October 2023.

\bibitem[Wahl et~al.(2025)Wahl, Rousselot, Draxler, and
  Koethe]{wahlTRADETransferDistributions2025}
Wahl, S., Rousselot, A., Draxler, F., and Koethe, U.
\newblock {{TRADE}}: {{Transfer}} of {{Distributions}} between {{External
  Conditions}} with {{Normalizing Flows}}.
\newblock In \emph{The 28th {{International Conference}} on {{Artificial
  Intelligence}} and {{Statistics}}}, February 2025.

\bibitem[Woo \& Ahn(2024)Woo and Ahn]{wooIteratedEnergybasedFlow2024}
Woo, D. and Ahn, S.
\newblock Iterated {{Energy-based Flow Matching}} for {{Sampling}} from
  {{Boltzmann Densities}}, August 2024.

\bibitem[Zhai et~al.(2024)Zhai, Zhang, Nakkiran, Berthelot, Gu, Zheng, Chen,
  Bautista, Jaitly, and Susskind]{zhaiNormalizingFlowsAre2024}
Zhai, S., Zhang, R., Nakkiran, P., Berthelot, D., Gu, J., Zheng, H., Chen, T.,
  Bautista, M.~A., Jaitly, N., and Susskind, J.
\newblock Normalizing {{Flows}} are {{Capable Generative Models}}, December
  2024.

\bibitem[Zhang et~al.(2023)Zhang, Chen, Liu, Courville, and
  Bengio]{zhangDiffusionGenerativeFlow2023a}
Zhang, D., Chen, R. T.~Q., Liu, C.-H., Courville, A., and Bengio, Y.
\newblock Diffusion {{Generative Flow Samplers}}: {{Improving}} learning
  signals through partial trajectory optimization.
\newblock In \emph{The {{Twelfth International Conference}} on {{Learning
  Representations}}}, October 2023.

\bibitem[Zhang \& Chen(2021)Zhang and Chen]{zhangPathIntegralSampler2021}
Zhang, Q. and Chen, Y.
\newblock Path {{Integral Sampler}}: {{A Stochastic Control Approach For
  Sampling}}.
\newblock In \emph{International {{Conference}} on {{Learning
  Representations}}}, October 2021.

\bibitem[Zheng et~al.(2024)Zheng, He, Liu, Shi, Lu, Feng, Ju, Wang, Zhu, Min,
  Zhang, Tang, Hao, Jin, Chen, No{\'e}, Liu, and
  Liu]{zhengPredictingEquilibriumDistributions2024}
Zheng, S., He, J., Liu, C., Shi, Y., Lu, Z., Feng, W., Ju, F., Wang, J., Zhu,
  J., Min, Y., Zhang, H., Tang, S., Hao, H., Jin, P., Chen, C., No{\'e}, F.,
  Liu, H., and Liu, T.-Y.
\newblock Predicting equilibrium distributions for molecular systems with deep
  learning.
\newblock \emph{Nat Mach Intell}, 6\penalty0 (5):\penalty0 558--567, May 2024.
\newblock ISSN 2522-5839.
\newblock \doi{10.1038/s42256-024-00837-3}.

\end{thebibliography}
\bibliographystyle{icml2025}

\newpage
\appendix
\onecolumn


\section{Internal Coordinate Representation} \label{SI:internal_coordinates}

As discussed in the main text, we use an internal coordinate representation based on bond lengths, angles, 
and dihedral angles to represent the conformations of the molecular systems.
As discussed in the next section, we use splines as the invertible transformations in our coupling blocks.
These splines are only defined for mappings from the interval $[0,B]$ to $[0,B]$. Therefore, we scale all
internal coordinates to fit in this range (here, $B=1$).

To achieve this, we divide all dihedral angles by $2\pi$. Furthermore, bond lengths and angles are transformed 
as 

\begin{align}
    \eta_i^\prime = (\eta_i - \eta_{i;\text{min}}) / \sigma + 0.5 \, .
\end{align}

Here, $\eta_\text{min}$ is the value of the corresponding degree of freedom from a minimum energy structure 
obtained from minimizing the initial structure with the force field. $\sigma$ was empirically chosen as \SI{0.07}{\nano \meter}
for bond length dimensions and \SI{0.5730}{\radian} for angle dimensions.

For the peptides studied in this work, two chiral forms (mirror images) exist. In nature,
one almost exclusively finds only one of the two (L-form). However, since the potential 
energy is invariant to the mirror symmetry, there is no preference given by the energy model itself.
Previous work \cite{midgleyFlowAnnealedImportance2023,schopmansConditionalNormalizingFlows2024} simply 
filtered the ``wrong'' R-chirality 
during training. In contrast, we directly constrain generation to the L-chirality by 
restricting the output bounds of the splines that generate the dihedrals of the hydrogens at the chiral centers to the range $[0.5, 1.0]$.
For this, we transform the corresponding dimensions as $\eta_i \rightarrow \eta_i \cdot 0.5 + 0.5$ after the flow generated them.
This entirely removes the R-chirality from the space the flow can generate.

Furthermore, there is no preference given for the permutation of hydrogens in \ce{CH3} groups.
However, since the ground truth molecular dynamics simulations start from a
given starting configuration, a preference does exist in the ground truth.
Therefore, we restrict the generated distribution of the flow to this
preference, by constraining the spline output range of the respective dihedral
angles, analogous to how we constrain the chirality (see above).

\section{Architecture} \label{SI:architecture}

\begin{figure*}[h]
\begin{center}
\vskip 0.2in
\centerline{\includegraphics{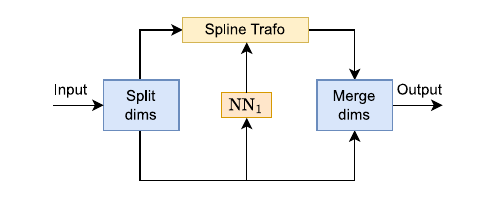}}
\caption{Illustration of a normalizing flow coupling layer.}
\label{SI:coupling_flow}
\end{center}
\vskip -0.2in
\end{figure*}

For the normalizing flow architecture, we use an architecture similar to previous
works \cite{midgleyFlowAnnealedImportance2023,schopmansConditionalNormalizingFlows2024}.
As the invertible transformation in the coupling layers, we use monotonic rational-quadratic splines \cite{durkanNeuralSplineFlows2019a}
that map the interval $[0,1]$ to $[0,1]$ using
monotonically increasing rational-quadratic functions with $K=8$ bins.
 
We use 8 pairs of neural spline coupling layers. In each pair, we use a randomly
generated mask to decide which dimensions to transform and which dimensions to
condition the transformation on (see Figure~\ref{SI:coupling_flow}). In the
second coupling of each pair, the inverted mask is used. The dimensions of the dihedral angles 
are treated using circular splines \cite{rezendeNormalizingFlowsTori2020a} to incorporate the correct topology. 
After each coupling layer, we add a random (but fixed) periodic shift to the dihedral angle dimensions.

The latent distribution $q_Z$ of the normalizing flow is a uniform distribution in the range $[0,1]$ for the dihedral angle dimensions,
and a Gaussian distribution with $\sigma=0.5$ and $\sigma=0.1$, truncated to the range $[0,1]$, for the bond length and angle dimensions.

As the conditioning network ($\text{NN}_1$ in Figure~\ref{SI:coupling_flow}), we
use a fully connected neural network with hidden dimensions
[256,256,256,256,256] and ReLU activation functions. Previous works used a fully connected neural network with
a skip connection
\cite{midgleyFlowAnnealedImportance2023,schopmansConditionalNormalizingFlows2024}, however, we found no benefit in this and therefore did not
use a skip connection. To incorporate their periodicity, dihedral angles are represented 
as $(\cos \psi, \sin \psi)^{\top}$ before being passed as input to the neural network.

We used the same architecture for all experiments of all methods. The number of
parameters in the architecture of each system can be found in
Table~\ref{SI:tab_parameter_count}.
All experiments used the Adam optimizer \cite{kingmaAdamMethodStochastic2017} to train the flow.

\begin{table}[h]
\caption{Number of parameters in the normalizing flow architecture for each system.}
\label{SI:tab_parameter_count}
\vskip 0.15in
\begin{center}
\begin{small}
\begin{sc}
\begin{tabular}{cccc}
\toprule
& Alanine Dipeptide & Alanine Tetrapeptide & Alanine Hexapeptide \\
\midrule
Number of parameters & \num{7421512} & \num{9452376} & \num{12124616} \\
\bottomrule
\end{tabular}
\end{sc}
\end{small}
\end{center}
\vskip -0.1in
\end{table}

To implement the normalizing flow models and the internal coordinate representations,
we used the \emph{bgflow} \cite{noeBgflow2024} and \emph{nflows} \cite{conordurkanNflowsNormalizingFlows2020} libraries with \emph{PyTorch} \cite{paszkePyTorchImperativeStyle2019}.

\section{Molecular Systems} \label{sec:SI:mol_systems}

\begin{table}[h]
\caption{Overview of the molecular systems. The number of constrained bonds is given in brackets.}
\label{SI:tab_molecular_systems}
\vskip 0.15in
\begin{center}
\begin{small}
\begin{sc}
\begin{tabular}{ccccccc}
\toprule
Name & Sequence & NO. atoms & NO. hydrogens & NO. bonds & NO. angles & NO. torsions \\
\midrule
\shortstack[c]{alanine \\ dipeptide}    & ACE-ALA-NME & 22 & 12 & 21 & 20 & 19 \\
\shortstack[c]{alanine \\ tetrapeptide} & ACE-3$\cdot$ALA-NME & 42 & 22 & 19 (+ 22) & 40 & 39 \\
\shortstack[c]{alanine \\ hexapeptide}  & ACE-5$\cdot$ALA-NME & 62 & 32 & 29 (+ 32) & 60 & 59 \\
\bottomrule
\end{tabular}
\end{sc}
\end{small}
\end{center}
\vskip -0.1in
\end{table}

To avoid diverging van der Waals energies due to atom clashes, we train with a
regularized energy function \cite{midgleyFlowAnnealedImportance2023}:

\begin{align}
&E_{\text{reg.}}(E) = \notag \\
&\begin{cases} 
E, & \text{if } E \leq E_{\text{high}}, \\
\log(E - E_{\text{high}} + 1) + E_{\text{high}}, & \text{if } E_{\text{high}} < E \leq E_{\text{max}}, \\
\log(E_\text{max} - E_{\text{high}} + 1) + E_{\text{high}}, & \text{if } E > E_{\text{max}}. \label{eq:energy_reg}
\end{cases}
\end{align}

For all systems, we used the energy regularization parameters $E_\text{high}=\num{1e8}$ and $E_\text{max}=\num{1e20}$.

Throughout this work, we use visualizations of the free energy $ F=-k_\text{B} T
\ln p(\phi_i, \psi_i) $ of the backbone dihedral angles of the systems (Ramachandran plots). For all Ramachandran plots, we used \num{1e7} samples if not otherwise specified. Furthermore, the axes
are in scaled internal coordinates in the range $[0,1]$.

\section{Force Field and Ground Truth Simulations}
All ground truth simulations have been performed with OpenMM 8.0.0 \cite{eastmanOpenMM8Molecular2024}
using the CUDA platform. Ground truth energy evaluations during training have
been performed with 18 workers in parallel using the OpenMM CPU Platform.

Details on the performed simulations and force field parameters for each system can be
found in Table \ref{SI:tab_ground_truth_simulations}. For all systems, we used 
variants of Amber force fields \cite{d.a.caseAmber20232023}. For alanine dipeptide, the
parameters are identical to the ones used in the FAB publication
\cite{midgleyFlowAnnealedImportance2023}. 
We use the dataset
made available by \citeauthor{stimperAlanineDipeptideImplicit2022} \yrcite{stimperAlanineDipeptideImplicit2022} as our test dataset.
In addition to this ground truth test dataset, we performed another molecular dynamics simulation for alanine dipeptide at
\SI{300}{\kelvin} (see Table \ref{SI:tab_ground_truth_simulations}). We used
\SI{50}{\nano \second} for equilibration and a production simulation time of
\SI{5}{\micro \second}. The small time step of \SI{1}{\femto \second} was chosen
since this system does not use hydrogen bond length constraints. This additional
simulation was used for training the forward KLD experiments and to create a separate validation
dataset.

The force field parameters of the alanine tetrapeptide system match those used
in the temperature steerable flow publication
\cite{dibakTemperatureSteerableFlows2022}. However, no public dataset for this
system was available, which is why we performed two replica exchange molecular
dynamics (REMD) simulations to obtain \SI{300}{\kelvin} ground truth
data. The hexapeptide system was, to the best of our knowledge, not used in
previous publications, so we also here performed REMD simulations to obtain a ground
truth dataset for evaluation. All REMD simulations used \SI{200}{\nano \second}
equilibration without exchanges, \SI{200}{\nano \second} equilibration with
exchanges, and \SI{1}{\micro \second} for the production simulation. For both
the tetrapeptide and hexapeptide, we used one simulation for the ground truth
test dataset and the other simulation to form the training dataset for the forward KLD 
experiments and to create a separate validation dataset.

Additionally to the simulations at \SI{300}{\kelvin}, we performed high-temperature simulations at \SI{1200}{\kelvin} to create validation datasets for the reverse KLD pre-training (see
Table \ref{SI:tab_ground_truth_simulations}).

All datasets were subsampled randomly from the total production MD trajectories. The ground truth test datasets at \SI{300}{\kelvin} contain \num{1e7} samples, the datasets used for the forward KLD experiments 
contain \num{1e6} samples. The additional validation datasets at \SI{300}{\kelvin} and \SI{1200}{\kelvin}
contain \num{1e6} samples.

\begin{table*}[h]
\caption{Overview of the molecular dynamics simulations performed to obtain the ground truth datasets. We only specify the production simulation time.
In the case of REMD simulations, this is the simulation time for each replica.}
\vskip 0.15in
\begin{center}
\begin{small}
\begin{sc}
\resizebox{\textwidth}{!}{\begin{tabular}{cccccc}
\toprule
System & Force field & Constraints & $T$ / \si{\kelvin} & Sim. time / \si{\micro \second} & Time step / \si{\femto \second} \\
\midrule

\multirow{3}{*}{\shortstack[c]{alanine \\ dipeptide}}  & \multirow{3}{*}{\shortstack[c]{AMBER ff96 \\ with OBC1 \\ implicit solvation}}  & \multirow{3}{*}{None} & \multirow{3}{*}{\shortstack[c]{300 \\ 1200}} & \multirow{3}{*}{\shortstack[c]{\num{5.0}}}  & \multirow{3}{*}{\shortstack[c]{\num{1.0}}} \\
\\
\\

\midrule

\multirow{3}{*}{\shortstack[c]{alanine \\ tetrapeptide}}  & \multirow{3}{*}{\shortstack[c]{AMBER99SB-ILDN \\ with AMBER99 OBC \\ implicit solvation}} & \multirow{3}{*}{\shortstack[c]{Hydrogen \\ bond lengths}} & \multirow{3}{*}{\shortstack[c]{300, 332, 368, 408, 451, 500 (REMD) \\ 1200}}  & \multirow{3}{*}{\shortstack[c]{\num{1.0} \\ \num{2.5}}}  & \multirow{3}{*}{\shortstack[c]{\num{2.0} \\ \num{1.0}}}   \\
\\
\\

\midrule

\multirow{3}{*}{\shortstack[c]{alanine \\ hexapeptide}}  & \multirow{3}{*}{\shortstack[c]{AMBER99SB-ILDN \\ with AMBER99 OBC \\ implicit solvation}} & \multirow{3}{*}{\shortstack[c]{Hydrogen \\ bond lengths}}  & \multirow{3}{*}{\shortstack[c]{300, 332, 368, 408, 451, 500 (REMD) \\ 1200}}   & \multirow{3}{*}{\shortstack[c]{\num{1.0} \\ \num{2.5}}} & \multirow{3}{*}{\shortstack[c]{\num{2.0} \\ \num{1.0}}}  \\
\\
\\

\bottomrule
\end{tabular}}
\end{sc}
\end{small}
\label{SI:tab_ground_truth_simulations}
\end{center}
\vskip -0.1in
\end{table*}

\section{Metrics Details} \label{SI:sec:metrics_details}
\paragraph{RAM KLD} To obtain comparable results to those in the original FAB publication \cite{midgleyFlowAnnealedImportance2023}, we
evaluated the forward KLD of the Ramachandran plots in the same way. First, we calculated the probability density 
of the Ramachandran plot of the ground truth and that of the flow distribution 
on a $100$x$100$ grid, using \num{1e7} samples for both. Then, the forward KLD is calculated between the two distributions.

\paragraph{RAM KLD W. RW} We repeat the same procedure to assess the obtained
Ramachandran plot after reweighting to the target distribution. Here, analogous
to \citeauthor{midgleyFlowAnnealedImportance2023}
\yrcite{midgleyFlowAnnealedImportance2023}, we clipped the \SI{0.01}{\percent}
highest importance weights to the lowest value among them. This is necessary
because of outliers in the importance weights due to flow numerics.

\paragraph{ESS} The ESS was calculated according to the following equation
\cite{midgleySE3EquivariantAugmented2023}:

\begin{align}
    &\frac{n_{\mathrm{e}, \mathrm{rv}}}{N}=\frac{1}{N \sum_{i=1}^N \bar{w}\left(x_i\right)^2} \label{eq:reverse_ESS} \\
    &\text { with } \quad x_i \sim q_X\left(x_i; \theta\right) \, \text{,} \quad \bar{w}(x_i) = \frac{w(x_i)}{\sum_{i=1}^N w(x_i)} \notag
\end{align}

Also here, we clipped the \SI{0.01}{\percent} highest importance weights to the
lowest value among them. Furthermore, the ESS was calculated with respect to the
regularized energy function (Equation~\ref{eq:energy_reg}).

While one can also calculate the forward ESS, which uses samples from the ground
truth \cite{midgleySE3EquivariantAugmented2023}, in practice we found this
metric to yield spurious results, depending heavily on the chosen importance
weight clipping value. Therefore, we chose to only use the reverse ESS, even
though it does not capture mode collapse.

\section{TA-BG}
\subsection{Workflow Variations}
As described in the main text, next to our buffered iterative annealing
workflow, we also tried training a temperature- conditioned normalizing flow on
the whole continuous temperature range. To fix the distribution at
\SI{1200}{\kelvin} to the distribution learned by the reverse KLD, we used a
split architecture to train with temperature-conditioning:

\begin{itemize}
    \item A base flow generates samples at \SI{1200}{\kelvin}. This was trained with the reverse KLD at \SI{1200}{\kelvin}, and the model parameters of this base were frozen afterwards.
    \item A temperature-conditioned head flow is added to the base flow, which transforms the high-temperature samples to lower temperatures. The spline couplings of the head flow are scaled in such a way that they always output the identity for $T=\SI{1200}{\kelvin}$.
\end{itemize}

In each batch, we sampled a support temperature $T_\text{support}$ and a target 
temperature $T_\text{reweight}$ and performed reweighted forward KLD training:

\begin{align}
    \mathcal{L}_\text{reweight}=-\mathop{\mathbb{E}}_{{x \sim q_X(x; T_\text{support};\theta)}} \underbrace{\frac{\exp(\frac{-E(x)}{k_\text{B} T_\text{reweight}})}{q_X(x; T_\text{support};\theta)}}_{\text{Stopped gradients}} \log q_X(x; T_\text{reweight};\theta) \label{SI:eq:alt_training_obj}
\end{align}

We also used this training objective with self-normalized importance weights within each batch, and with resampling each batch according to the importance weights.

The training objective in Equation \ref{SI:eq:alt_training_obj} is similar to the one introduced by \citeauthor{wahlTRADETransferDistributions2025} \yrcite{wahlTRADETransferDistributions2025}, but does not require the estimate of the partition function. In practice, we found the iterative annealing workflow to yield more accurate results, while also being more sampling efficient compared to using Equation~\ref{SI:eq:alt_training_obj}.

A systematic comparison of using Equation~\ref{SI:eq:alt_training_obj} and using the training objective introduced by \citeauthor{wahlTRADETransferDistributions2025} \yrcite{wahlTRADETransferDistributions2025} can be explored in future work.


\subsection{Hyperparameters of Main Experiments}

\begin{table}[h]
\caption{Hyperparameters of the TA-BG experiments, annealing from
\SI{1200}{\kelvin} to \SI{300}{\kelvin}. The cosine annealing learning rate
scheduler is applied within each annealing iteration, so the learning rate
resets in the beginning of each new annealing iteration. ``Annealing iteration``
here also refers to the fine-tuning iterations with $T_{i+1}=T_i$.}
\label{SI:tab_hyper_annealing}
\vskip 0.15in
\begin{center}
\begin{small}
\begin{sc}
\begin{tabular}{>{\centering\arraybackslash}m{4cm}ccc}
\toprule
                    &   Dipeptide & Tetrapeptide & Hexapeptide  \\
\midrule
gradient descent steps per annealing iteration & \num{30000} & \num{20000} & \num{20000} \\
learning rate          & \num{5e-6} & \num{1e-5} & \num{5e-6} \\
batch size             & 2048 & 4096 & 2048 \\
LR scheduler           & cosine annealing & - & - \\
buffer samples drawn per annealing iteration & \num{5e6} & \num{5e6} & \num{1e7} \\
buffer resampled to & \num{2e6} & \num{2e6} & \num{2e6} \\
\bottomrule
\end{tabular}
\end{sc}
\end{small}
\end{center}
\vskip -0.1in
\end{table}

In all TA-BG experiments, we used the following annealing iterations:

\begin{align}
    &\underbrace{\SI{1200}{\kelvin}}_{T_1} \rightarrow \underbrace{\SI{1028.69}{\kelvin}}_{T_2} \rightarrow \SI{881.84}{\kelvin} \rightarrow \SI{755.95}{\kelvin}
    \rightarrow \SI{648.04}{\kelvin} \rightarrow \SI{555.52}{\kelvin} \notag \\ 
    &\rightarrow \SI{476.22}{\kelvin} \rightarrow \SI{408.24}{\kelvin}
    \rightarrow \SI{349.96}{\kelvin} \rightarrow \underbrace{\SI{300.0}{\kelvin}}_{T_{K-1}} \rightarrow \underbrace{\SI{300.0}{\kelvin}}_{T_K=T_\text{target}}
\end{align}

As described in the main text, the hexapeptide additionally used intermediate
fine-tuning iterations after each annealing iteration, not only in the very end:

\begin{align}
    &\underbrace{\SI{1200}{\kelvin}}_{T_1} \rightarrow \underbrace{\SI{1028.69}{\kelvin}}_{T_2} \rightarrow \SI{1028.69}{\kelvin} \rightarrow \SI{881.84}{\kelvin} \rightarrow \SI{881.84}{\kelvin} \rightarrow \SI{755.95}{\kelvin} \rightarrow \SI{755.95}{\kelvin} \notag \\
    &\rightarrow \SI{648.04}{\kelvin} \rightarrow \SI{648.04}{\kelvin} \rightarrow \SI{555.52}{\kelvin} \rightarrow \SI{555.52}{\kelvin} \rightarrow \SI{476.22}{\kelvin} \rightarrow \SI{476.22}{\kelvin} \rightarrow \SI{408.24}{\kelvin} \rightarrow \SI{408.24}{\kelvin} \notag \\
    &\rightarrow \SI{349.96}{\kelvin} \rightarrow \SI{349.96}{\kelvin} \rightarrow \underbrace{\SI{300.0}{\kelvin}}_{T_{K-1}} \rightarrow \underbrace{\SI{300.0}{\kelvin}}_{T_K=T_\text{target}} \label{SI:eq:hexa_temp_schedule}
\end{align}

As described in the main text, in each iteration of the annealing workflow, we
use a buffered dataset for training, resampled according to the importance
weights. Similarly to how we
clipped the importance weights to calculate the forward KLD of the reweighted
Ramachandrans, also here we clipped the highest \SI{0.01}{\percent} of the
importance weights to the lowest value among them. This can prevent
overemphasizing outliers in the importance weights, though the effect is
minimal.

We further note that instead of using only the training buffer dataset of the
current annealing iteration, one can also reuse older samples that correspond to
previous annealing iterations by reweighting them to the current target
temperature. To simplify the workflow, we did not pursue this option in our
experiments, but it might further increase sample efficiency.

\subsection{Ablation Experiments} \label{SI:sec:ta_bg_ablations} In this
section, we discuss several ablation experiments to analyze the impact of
different hyperparameter choices. To keep the computational cost down, we
performed each ablation experiment only once (except for the ablation
experiments on starting temperatures). All ablation experiments start from the
hyperparameters chosen in the main experiments, changing only the
hyperparameters specified.

When ablating choices in the annealing workflow, we start all experiments from
the same checkpoint pre-trained with reverse KLD at high temperature, to remove
variations in the pre-training.

\begin{figure*}[h]
\begin{center}
\vskip 0.2in
\centerline{\includegraphics{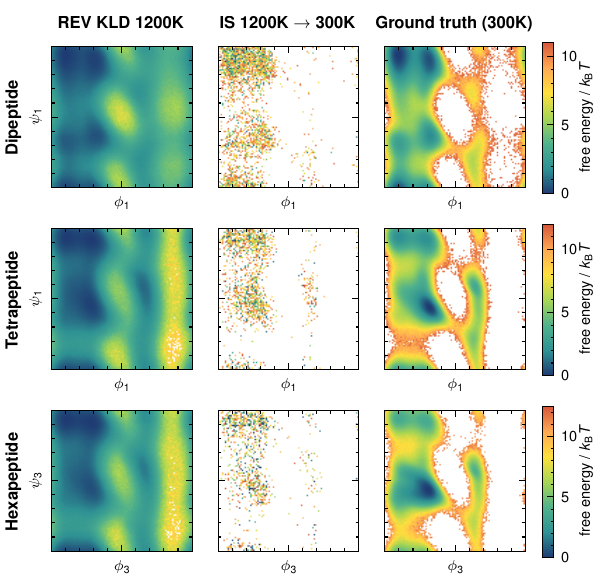}}
\caption{Free energy $ F=-k_\text{B} T \ln p(\phi_i, \psi_i) $ of 
dihedral angles (Ramachandran plots), reweighted directly from
\SI{1200}{\kelvin} to \SI{300}{\kelvin}. We used \num{1e6} samples for
importance sampling (middle).}
\label{SI:fig_direct_importance_sampling}
\end{center}
\vskip -0.2in
\end{figure*}

\paragraph{Direct Importance Sampling}
Figure~\ref{SI:fig_direct_importance_sampling} shows the result when directly
performing importance sampling from \SI{1200}{\kelvin} to \SI{300}{\kelvin},
without using the iterative annealing workflow. As one can see, the overlap
between the two distributions is very small, resulting in a very noisy result
due to the small effective sample size. This shows that a multistep annealing
workflow is necessary.

\paragraph{Starting Temperature}
As discussed in the main text, we perform high-temperature reverse KLD training
in the beginning of our workflow to pre-train the flow distribution. The
starting temperature $T_1$ at which reverse KLD training is performed needs to
be chosen high enough such that minima are less separated and mode collapse is
avoided.

\begin{figure*}[h]
\begin{center}
\vskip 0.2in
\centerline{\includegraphics{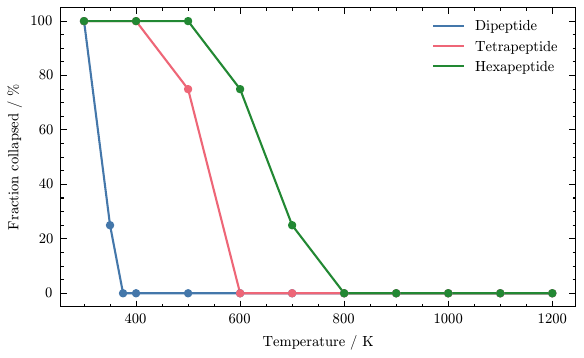}}
\caption{Ablation of starting temperature $T_1$: Fraction of reverse KLD experiments with mode collapse at a given temperature.}
\label{SI:fig_mode_collapse_study}
\end{center}
\vskip -0.2in
\end{figure*}

In Figure~\ref{SI:fig_mode_collapse_study}, we analyze the propensity for mode
collapse for each of the three studied systems. For each temperature, we
performed four reverse KLD experiments and display the fraction of experiments
with mode collapse among them (mode collapse is defined here by manual visual
inspection of the Ramachandran plots). As one can see, for each system, a
critical temperature exists above which mode collapse does not occur.

We note that the propensity for mode collapse not only depends on the
temperature, but also on the chosen batch size and the number of gradient descent
steps. The experiments in Figure~\ref{SI:fig_mode_collapse_study} were performed
with a batch size of \num{1024} and \num{250000} gradient descent steps.
Generally, when increasing the target temperature beyond the so-obtained
critical temperature, one can use a smaller batch size and fewer gradient
descent steps without obtaining mode collapse. This allows more efficient
training in terms of target energy evaluations. For our main experiments, we
thus chose a relatively large initial temperature $T_1=\SI{1200}{\kelvin}$. A
tradeoff exists: Increasing $T_1$ allows cheaper reverse KLD training without
mode collapse, but also requires more annealing steps to reach the desired
target temperature.

\begin{figure*}[h]
\begin{center}
\vskip 0.2in
\centerline{\includegraphics{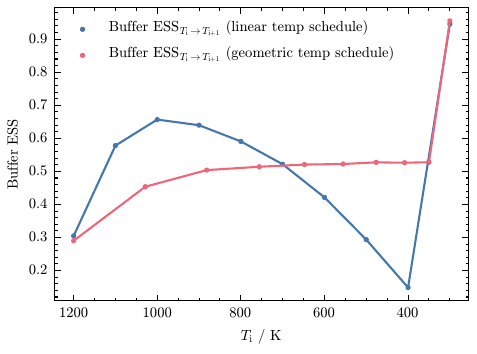}}
\caption{The ESS of the training buffer datasets $\mathcal{W}$ used to anneal
from $T_i$ to $T_{i+1}$, comparing a linear and geometric temperature schedule
for alanine dipeptide. The increase of the buffer ESS in the end is due to the
final fine-tuning iteration with $T_{i+1}=T_i$.}
\label{SI:fig_temp_schedule_aldp}
\end{center}
\vskip -0.2in
\end{figure*}

\begin{figure*}[h]
\begin{center}
\vskip 0.2in
\centerline{\includegraphics{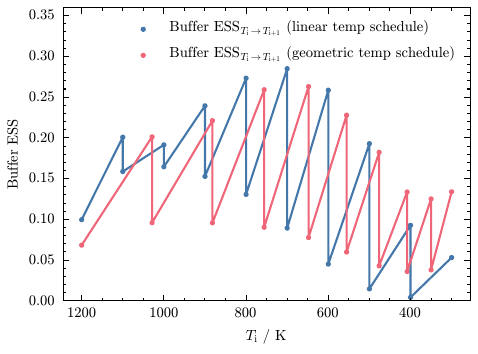}}
\caption{The ESS of the training buffer datasets $\mathcal{W}$ used to anneal
from $T_i$ to $T_{i+1}$, comparing a linear and geometric temperature schedule
for alanine hexapeptide. Note that the hexapeptide system includes intermediate
fine-tuning iterations ($T_{i+1}=T_{i}$), where the buffer ESS is increased due
to better overlap compared to the iterations where the temperature is lowered
($T_{i+1} < T_i$).}
\label{SI:fig_temp_schedule_hexa}
\end{center}
\vskip -0.2in
\end{figure*}

\paragraph{Temperature Schedule}
As discussed in Section~\ref{sec:methods}, we used a geometric progression
between $T_\text{start}$ and $T_\text{target}$ as the temperature schedule for
our main experiments. We compare this choice with a linear temperature schedule
for alanine dipeptide in Figure~\ref{SI:fig_temp_schedule_aldp}, and for the
hexapeptide in Figure~\ref{SI:fig_temp_schedule_hexa}. As one can see, the
geometric temperature schedule is able to keep an approximately constant buffer
ESS and therefore overlap between two consecutive distributions, while the
buffer ESS drops down significantly for the linear schedule. In the future, also
adaptive schedules can be explored
\cite{goshtasbpourAdaptiveAnnealedImportance2023} to further improve the
transition from one temperature level to the next.

\paragraph{Final Fine-Tuning}
As described in Section~\ref{sec:methods}, all TA-BG experiments used a final
fine-tuning iteration with $T_{K-1}=T_K$. Importance sampling to the same
temperature (without lowering the target temperature) yields higher overlap
between the two distributions, as is evident from the increased buffer ESS in
the final iteration shown in Figure~\ref{SI:fig_temp_schedule_aldp}. We found
empirically that such a final fine-tuning iteration with increased buffer ESS is
helpful to improve the final metrics for all systems, as is summarized in
Table~\ref{tab:SI_final_finetuning}.

\begin{figure*}[h]
\begin{center}
\vskip 0.2in
\centerline{\includegraphics{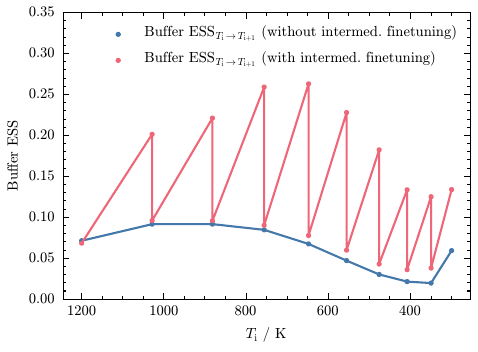}}
\caption{The ESS of the training buffer datasets $\mathcal{W}$ used to anneal
from $T_i$ to $T_{i+1}$, comparing the case with and without intermediate
fine-tuning iterations for the hexapeptide system.}
\label{SI:fig_intermed_finetuning_hexa}
\end{center}
\vskip -0.2in
\end{figure*}

\paragraph{Intermediate Fine-Tuning}
For the hexapeptide system, we additionally added intermediate fine-tuning
iterations after each annealing iteration, instead of only in the very end. We
compare the impact of such intermediate fine-tuning for all systems in
Table~\ref{tab:SI_intermed_finetuning}. As we can see, the impact is the largest
for the hexapeptide system, where including intermediate fine-tuning iterations
significantly boosts the final ESS (see also
Figure~\ref{SI:fig_intermed_finetuning_hexa}). Since such intermediate
fine-tuning costs additional target energy evaluations, we did not include them
for the two smaller systems.

\paragraph{NO. Annealing Iterations}
For our main experiments, we annealed the temperature 9 times for all systems.
We summarize the impact of the number of annealing iterations on the final
metrics for alanine dipeptide in Table~\ref{tab:SI_NO_annealing_steps}. Choosing
the number of annealing iterations is a tradeoff between improving the final
metrics and lowering the number of target evaluations.

\paragraph{NO. Samples Per Annealing Iteration}
Also the number of samples drawn to form the training buffer datasets in each
annealing iteration forms a tradeoff between improving the final metrics and
lowering the number of target evaluations. This is summarized in
Table~\ref{tab:SI_NO_samples} for alanine dipeptide, where we draw the specified
number of samples and resample this dataset to the same size for training in each
annealing iteration.

\begin{table*}[h]
\caption{Impact of final fine-tuning step on final metrics per system.}
\label{tab:SI_final_finetuning}
\vskip 0.15in
\begin{center}
\begin{small}
\begin{sc}
\begin{tabular}{ccccc}
\toprule
System & Intermed. Fine-tuning & Final Fine-tuning & NLL & ESS / \% \\
\midrule
\multirow{2}{*}{\shortstack[c]{alanine \\ dipeptide}} & Without & With & -213.667 & 95.85 \\ 
                                                                  & Without & Without & -213.666 & 95.55 \\ 
\multirow{2}{*}{\shortstack[c]{alanine \\ tetrapeptide}} & Without & With & -330.128 & 62.57 \\ 
                                                                  & Without & Without & -330.087 & 57.57 \\ 
\multirow{2}{*}{\shortstack[c]{alanine \\ hexapeptide}} & With & With & -501.510 & 15.38 \\ 
                                                                  & With & Without & -501.461 & 13.37 \\ 
\bottomrule
\end{tabular}
\end{sc}
\end{small}
\end{center}
\vskip -0.1in
\end{table*}

\begin{table*}[h]
\caption{Impact of intermediate fine-tuning on final metrics per system.}
\label{tab:SI_intermed_finetuning}
\vskip 0.15in
\begin{center}
\begin{small}
\begin{sc}
\begin{tabular}{cccc}
\toprule
System & Intermed. Fine-tuning & NLL & ESS / \% \\
\midrule
\multirow{2}{*}{\shortstack[c]{alanine \\ dipeptide}} & With & -213.671 & 96.46 \\ 
                                                                  & Without & -213.667 & 95.85 \\ 
\multirow{2}{*}{\shortstack[c]{alanine \\ tetrapeptide}} & With & -330.176 & 68.93 \\ 
                                                                  & Without & -330.128 & 62.57 \\ 
\multirow{2}{*}{\shortstack[c]{alanine \\ hexapeptide}} & With & -501.510 & 15.38 \\ 
                                                                  & Without & -499.580 & 7.56 \\ 
\bottomrule
\end{tabular}
\end{sc}
\end{small}
\end{center}
\vskip -0.1in
\end{table*}

\begin{table*}[h]
\caption{Impact of number of annealing iterations on final metrics for alanine dipeptide.}
\label{tab:SI_NO_annealing_steps}
\vskip 0.15in
\begin{center}
\begin{small}
\begin{sc}
\begin{tabular}{ccc}
\toprule
NO. annealing iterations & NLL & ESS / \% \\
\midrule

3 & -213.647 & 92.27 \\ 
5 & -213.663 & 95.06 \\ 
7 & -213.666 & 95.57 \\ 
9 & -213.667 & 95.85 \\ 
11 & -213.668 & 96.05 \\ 

\bottomrule
\end{tabular}
\end{sc}
\end{small}
\end{center}
\vskip -0.1in
\end{table*}

\begin{table*}[h]
\caption{Impact of number of samples drawn per annealing iteration on final metrics for alanine dipeptide.}
\label{tab:SI_NO_samples}
\vskip 0.15in
\begin{center}
\begin{small}
\begin{sc}
\begin{tabular}{ccc}
\toprule
NO. samples & NLL & ESS / \% \\
\midrule

\num{500000} & -213.556 & 84.49 \\ 
\num{1000000} & -213.649 & 92.47 \\ 
\num{2000000} & -213.663 & 95.12 \\ 
\num{5000000} & -213.670 & 96.43 \\ 
\num{10000000} & -213.673 & 96.86 \\ 

\bottomrule
\end{tabular}
\end{sc}
\end{small}
\end{center}
\vskip -0.1in
\end{table*}

\clearpage

\subsection{Scaling to Higher Dimensions}
Our proposed annealing workflow relies on importance sampling (IS) to anneal the
distribution iteratively from $T_i$ to $T_{i+1}$. Since importance sampling does
not scale well when increasing the dimensionality $N$ of the problem, we want to
shortly discuss how annealed importance sampling (AIS) \cite{nealAnnealedImportanceSampling2001} can be used in
place of IS in the future, and how this can mitigate potential problems
encountered with IS when scaling to larger systems.

The main problem of IS is that small local bias, e.g., the possibility of two
atoms clashing locally in the proposal, gets amplified due to the dimensionality
of the problem. We consider a molecular system with $N$ independent
neighborhoods, each with a chance $\eta$ for a clash under the proposal
distribution. A sample's importance weight is $1$ if no clashes occurred, and
$0$ otherwise. The ESS is $\text{ESS}=\frac{\left(\sum_{i=1}^M
w_i\right)^2}{M \sum_{i=1}^M w_i^2}=\frac{1}{M} \sum_{i=1}^M w_i$, so $\mathbb{E}(\text{ESS})
= (1-\eta)^N$. Thus, the ESS drops exponentially, which captures the
scaling problem of IS.

To demonstrate how AIS addresses this, we move to a continuous model (a similar
example and analysis can be found in \cite{midgleyFlowAnnealedImportance2023}):
The proposal distribution is an $N$-dimensional Gaussian with $\sigma=1.1$,
$\mu=0$, and the target distribution is an $N$-dimensional Gaussian with
$\sigma=1.0$, $\mu=0$. This is a simplified model of a single annealing
iteration. To perform AIS in this toy example, we use a single step of HMC with
$5$ leap-frog steps to transition between two intermediate distributions. We
further scale the number of intermediate distributions $T$ linearly with the
dimensionality $N$, we chose $T=5 \cdot N$ in this example. 

\begin{figure*}[h]
\begin{center}
\vskip 0.2in \centerline{\includegraphics{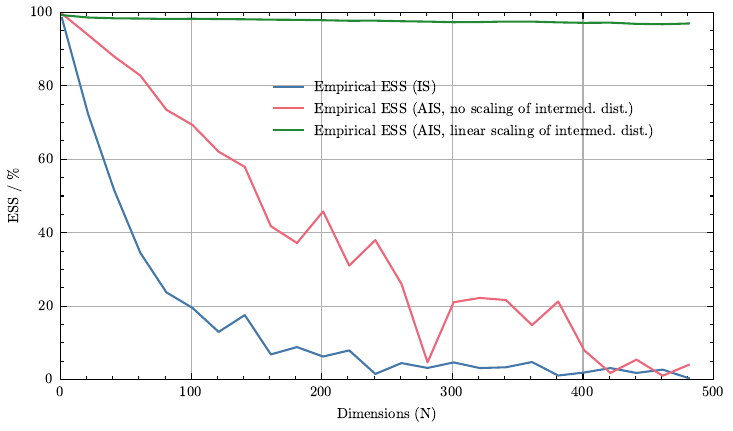}}
\caption{Comparison of the empirical ESS as a function of the number of
dimensions $N$ for a Gaussian toy system, as obtained using vanilla IS, AIS
without increasing the number of intermediate distributions, and AIS when
linearly scaling up the number of intermediate distributions with $N$.}
\label{SI:fig_AIS_IS_comparison}
\end{center}
\vskip -0.2in
\end{figure*}

We visualize the results of this toy example in
Figure~\ref{SI:fig_AIS_IS_comparison}. As one can see, the ESS of vanilla IS
drops exponentially. However, when using AIS and scaling the number of
intermediate distributions linearly with $N$, one can obtain an approximately
constant ESS. While this is, of course, a simplified setup, it still captures
the essence of why IS fails for higher dimensions and how AIS can mitigate this
issue. We also refer to \cite{nealAnnealedImportanceSampling2001} for a theoretical analysis. 
While for molecular systems, exact factorizability of the distribution is
of course not given, AIS can still remove atom clashes, etc., which are often
local effects. Thus, by using AIS instead of IS, potentially arising scaling
problems can be avoided in the future.

\section{FAB}
\subsection{Hyperparameters of Main Experiments}

For the FAB experiments, we started from the hyperparameters reported in the
original publication. To make the original FAB hyperparameters a good starting
point, we scaled our internal coordinates from the range of the splines [0,1] to
[0,10], which is the range used in the original FAB implementation. With this, we used an
initial HMC step size of $0.05$ for all experiments. Since FAB obtained
significantly better results by using a prioritized replay buffer, and we were
able to reproduce this finding, we chose the same replay buffer used originally
by FAB for all experiments.

\begin{table}[h]
\caption{Hyperparameters of the FAB experiments at \SI{300}{\kelvin}. All experiments 
used a cosine annealing learning rate scheduler with a single cycle.}
\label{SI:tab_hyper_fab}
\vskip 0.15in
\begin{center}
\begin{small}
\begin{sc}
\begin{tabular}{>{\centering\arraybackslash}m{4cm}ccc}
\toprule
                    &   Dipeptide & Tetrapeptide & Hexapeptide  \\
\midrule
gradient descent steps & \num{50000} & \num{50000} & \num{50000} \\
learning rate          & \num{1e-4} & \num{1e-4} & \num{1e-4} \\
batch size             & 1024 & 1024 & 1024 \\
grad norm clipping     & 1000.0 & 1000.0 & 1000.0 \\
lr linear warmup steps & 1000 & 1000 & 1000 \\
weight decay (L2)      & \num{1e-5} & \num{1e-5} & \num{1e-5} \\
NO. intermed. dist.    & 8 & 8 & 8 \\
NO. inner hmc steps    & 4 & 4 & 8 \\
\bottomrule
\end{tabular}
\end{sc}
\end{small}
\end{center}
\vskip -0.1in
\end{table}

\subsection{Ablation Experiments}
As with most sampling approaches, the number of target evaluations and the
accuracy of the obtained distribution form a tradeoff. For the hexapeptide, FAB
was not able to resolve the metastable high-energy region accurately. Therefore,
we performed additional experiments where we varied the number of intermediate
distributions and the number of HMC steps. This improves the results slightly,
while requiring significantly more target energy evaluations (see
Table~\ref{SI:FAB_variations}).

For the smaller systems, alanine dipeptide and alanine tetrapeptide, we further
performed experiments with smaller batch sizes while using the same number of
gradient descent steps. This lowers the number of required target energy
evaluations. However, as one can see from Table~\ref{SI:FAB_variations}, this
comes at the cost of further increasing the NLL.

\begin{table}[h]
\caption{Impact of hyperparameter choices on the NLL of FAB. In bold are the hyperparameters that were used for the main experiments,
chosen as a trade-off between computational cost and accuracy.
Only the main experiments were performed four times, all other experiments were performed once.}
\label{SI:FAB_variations}
\vskip 0.15in
\begin{center}
\begin{small}
\begin{sc}
\begin{tabular}{cccccccc}
\toprule
\multirow{2}{*}{System} & \multirow{2}{*}{batch size} & NO. intermed. & NO. inner & \multirow{2}{*}{PE EVALS $\downarrow$} & \multirow{2}{*}{NLL $\downarrow$} \\
       &            & dist.         & HMC steps &          &                   &                        \\
\midrule
\multirow{3}{*}{\shortstack[c]{alanine \\ dipeptide}}    & \textbf{1024} & \textbf{8} & \textbf{4} & \textbf{\num{2.13e8}} & \textbf{\num{-213.653(000)}} \\
                                                                     & 512              & 8 & 4 & \num{1.07e8} & \num{-213.643} \\ 
                                                                     & 256              & 8 & 4 & \num{5.33e7} & \num{-213.623} \\ 
\midrule
\multirow{3}{*}{\shortstack[c]{alanine \\ tetrapeptide}} & \textbf{1024} & \textbf{8} & \textbf{4} & \textbf{\num{2.13e8}} & \textbf{\num{-330.100(002)}} \\
                                                                     & 512              & 8 & 4 & \num{1.07e8} & \num{-330.019} \\ 
                                                                     & 256              & 8 & 4 & \num{5.33e7} & \num{-329.874} \\ 
\midrule
\multirow{5}{*}{\shortstack[c]{alanine \\ hexapeptide}} & 1024              & 8  & 4 & \num{2.13e8} & \num{-501.157} \\ 
                                                                     & 512              & 8  & 4 & \num{1.07e8} & \num{-500.857} \\ 
                                                                     & 1024             & 16 & 4 & \num{4.20e8} & \num{-501.255} \\ 
                                                                     & \textbf{1024} & \textbf{8}  & \textbf{8} & \textbf{\num{4.20e8}} & \textbf{\num{-501.268(0.008)}} \\
                                                                     & 1024             & 16 & 8 & \num{8.34e8} & \num{-501.327} \\ 
\bottomrule
\end{tabular}
\end{sc}
\end{small}
\end{center}
\vskip -0.1in
\end{table}

\newpage
\section{Reverse KLD}
\subsection{Hyperparameters of Main Experiments}

\begin{table}[h]
\caption{Hyperparameters of the reverse KLD experiments. This includes the \SI{300}{\kelvin} experiments, but also
the \SI{1200}{\kelvin} experiments used to start the temperature-annealing workflow. All experiments 
used a cosine annealing learning rate scheduler with a single cycle.}
\label{SI:tab_hyper_rev_kld}
\vskip 0.15in
\begin{center}
\begin{small}
\begin{sc}
\begin{tabular}{>{\centering\arraybackslash}m{3cm}cccccc}
\toprule
                    &   Dipeptide & Tetrapeptide & Hexapeptide  & Dipeptide & Tetrapeptide & Hexapeptide \\
   $T$                 & \SI{300}{\kelvin} & \SI{300}{\kelvin} & \SI{300}{\kelvin} & \SI{1200}{\kelvin} & \SI{1200}{\kelvin} & \SI{1200}{\kelvin} \\
\midrule
gradient descent steps & \num{250000} & \num{250000} & \num{250000} & \num{100000} & \num{100000} & \num{250000} \\
learning rate          & \num{1e-4} & \num{1e-4} & \num{1e-4} & \num{1e-4} & \num{1e-4} & \num{1e-4} \\
batch size             & 1024 & 1024 & 1024 & 256 & 256 & 512 \\
grad norm clipping     & 100.0 & 100.0 & 100.0 & 100.0 & 100.0 & 100.0 \\
lr linear warmup steps & 1000 & 1000 & 1000 & 1000 & 1000 & 1000 \\
weight decay (L2)      & \num{1e-5} & \num{1e-5} & \num{1e-5} & \num{1e-5} & \num{1e-5} & \num{1e-5} \\
NO. highest energy values removed & 40 & 40 & 40 & 10 & 10 & 20 \\
\bottomrule
\end{tabular}
\end{sc}
\end{small}
\end{center}
\vskip -0.1in
\end{table}

\section{Forward KLD}
\subsection{Hyperparameters of Main Experiments}

\begin{table}[h]
\caption{Hyperparameters of the forward KLD experiments at \SI{300}{\kelvin}. All experiments 
used a cosine annealing learning rate scheduler with a single cycle.}
\label{SI:tab_hyper_fwd_kl}
\vskip 0.15in
\begin{center}
\begin{small}
\begin{sc}
\begin{tabular}{>{\centering\arraybackslash}m{4cm}ccc}
\toprule
                    &   Dipeptide & Tetrapeptide & Hexapeptide  \\
\midrule
gradient descent steps & \num{100000} & \num{100000} & \num{120000} \\
learning rate          & \num{5e-5} & \num{5e-5} & \num{5e-5} \\
batch size             & 1024 & 1024 & 1024 \\
\bottomrule
\end{tabular}
\end{sc}
\end{small}
\end{center}
\vskip -0.1in
\end{table}

\clearpage

\section{Comparison of Computational Cost} \label{SI:sec_comp_cost} 

In this section, we compare the computational cost of TA-BG and FAB, summarized
in Table~\ref{SI:tab_comp_cost}. We point out that neither of the two
implementations was optimized for speed, thus the wall times are only
approximately indicative of the actually achievable performance.

As already discussed, TA-BG is more efficient compared to FAB in terms of target
energy evaluation. However, this increased sampling efficiency comes with the
cost of more flow evaluations, since we train for an extended period of time on
large buffer datasets. The force field evaluations of the benchmark systems in
this work are relatively inexpensive. Thus, the total wall time of FAB is
currently slightly lower compared to TA-BG for the systems investigated here.

However, we note that while the force field evaluations are inexpensive, they
are also not very accurate. When applying our approach to systems with more
expensive target evaluations, such as ones based on foundation model force
fields or density functional theory, target evaluations become dominant,
favoring methods with improved sampling efficiency, such as TA-BG.

\begin{table}[h]
\caption{Computational cost of TA-BG and FAB, comparing the number of target
evaluations, the number of flow evaluations (batched), and the total wall time
(excluding evaluation). To determine the wall times, we ran 4 experiments in
parallel on a compute node with 2 $\times$ AMD EPYC Rome 7402 CPU (24 cores each) and 4
$\times$ NVIDIA A100 GPU.}
\label{SI:tab_comp_cost}
\vskip 0.15in
\begin{center}
\begin{small}
\begin{sc}
\resizebox{\textwidth}{!}{\begin{tabular}{ccccc}
\toprule
System & Method & NO. Target Evals & NO. Flow Evals (batches) & Wall Time \\
\midrule
\multirow{7}{*}{\shortstack[c]{alanine \\ dipeptide}}    & FAB & \num{2.13e8} & \num{2.06e5} (AIS update buffer, bs 1024) & \SI{18.0}{\hour} \\
                                                         &     &              & + \num{5e4} (training, bs 1024) \\
                                                         &     &              & = \num{2.56e5} \\
                                                         & TA-BG & \num{2.56e7} (rev. KLD pre-training) & \num{1e5} (rev. KLD pre-training, bs 256)         & \SI{22.2}{\hour} \\
                                                         &       & + \num{5e7} (annealing)              & + \num{5e7} / 4096 (buffer creation, bs 4096) \\
                                                         &       & = \num{7.56e7}                       & + \num{3e5} (fwd. KLD annealing, bs 2048) \\
                                                         &       &                                      & = \num{4.12e5} \\
\midrule
\multirow{7}{*}{\shortstack[c]{alanine \\ tetrapeptide}} & FAB & \num{2.13e8} & \num{2.06e5} (AIS update buffer, bs 1024) & \SI{19.67}{\hour} \\
                                                         &     &              & + \num{5e4} (training, bs 1024) \\
                                                         &     &              & = \num{2.56e5} \\
                                                         & TA-BG & \num{2.56e7} (rev. KLD pre-training) & \num{1e5} (rev. KLD pre-training, bs 256)         & \SI{23.3}{\hour} \\
                                                         &       & + \num{5e7} (annealing)              & + \num{5e7} / 4096 (buffer creation, bs 4096) \\
                                                         &       & = \num{7.56e7}                       & + \num{2e5} (fwd. KLD annealing, bs 4096) \\
                                                         &       &                                      & = \num{3.12e5} \\
\midrule
\multirow{7}{*}{\shortstack[c]{alanine \\ hexapeptide}}  & FAB & \num{4.20e8} & \num{4.06e5} (AIS update buffer, bs 1024) & \SI{41.4}{\hour} \\
                                                         &     &              & + \num{5e4} (training, bs 1024) \\
                                                         &     &              & = \num{4.56e5} \\
                                                         & TA-BG & \num{1.28e8} (rev. KLD pre-training)   & \num{2.5e5} (rev. KLD pre-training, bs 512)         & \SI{52.2}{\hour} \\
                                                         &       & + \num{1.8e8} (annealing)              & + \num{1.8e8} / 4096 (buffer creation, bs 4096) \\
                                                         &       & = \num{3.08e8}                         & + \num{3.6e5} (fwd. KLD annealing, bs 2048) \\
                                                         &       &                                        & = \num{6.5e5} \\
\bottomrule
\end{tabular}}
\end{sc}
\end{small}
\end{center}
\vskip -0.1in
\end{table}

\clearpage
\section{2D GMM System}
\label{SI:sec_gmm}

\begin{figure*}[h]
\begin{center}
\vskip 0.2in \centerline{\includegraphics{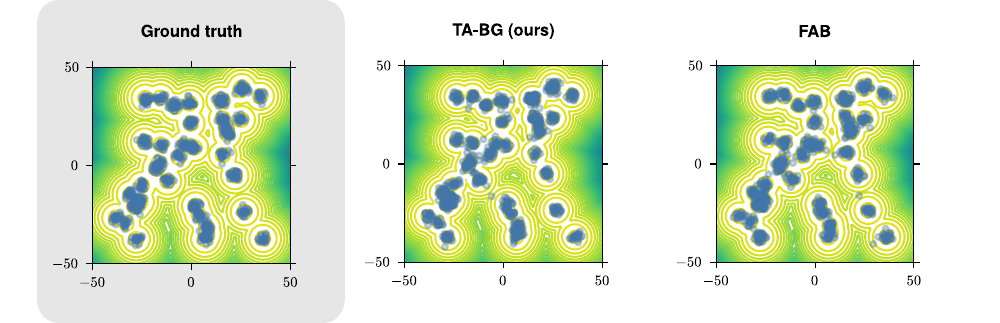}}
\caption{Comparison of obtained distributions for the 2D GMM system. We
visualize $1000$ samples from the ground truth distribution, TA-BG, and FAB. We
additionally visualize the contour lines of the ground truth log probability.}
\label{SI:fig_GMM}
\end{center}
\vskip -0.2in
\end{figure*}

Next to the molecular systems, we additionally consider the 40 Gaussian
mixture density in 2 dimensions as introduced by
\citeauthor{midgleyFlowAnnealedImportance2023}
\yrcite{midgleyFlowAnnealedImportance2023}. The ground truth distribution, as
well as the results obtained with TA-BG and FAB, are visualized in
Figure~\ref{SI:fig_GMM}.

We report the obtained NLL and ESS in Table~\ref{tab:SI_GMM_results}. This table
additionally contains results for diffusion-based samplers: Path Integral
Sampler (PIS, \citeauthor{zhangPathIntegralSampler2021}
\yrcite{zhangPathIntegralSampler2021}), Denoising Diffusion Sampler (DDS,
\citeauthor{vargasDenoisingDiffusionSamplers2022}
\yrcite{vargasDenoisingDiffusionSamplers2022}), and Iterated Denoising Energy
Matching (iDEM, \citeauthor{akhound-sadeghIteratedDenoisingEnergy2024a}
\yrcite{akhound-sadeghIteratedDenoisingEnergy2024a}). For these methods, we show
the results reported by
\citeauthor{akhound-sadeghIteratedDenoisingEnergy2024a}
\yrcite{akhound-sadeghIteratedDenoisingEnergy2024a}.

FAB (REAL NVP) in Table~\ref{tab:SI_GMM_results} refers to the original FAB
results reported by
\citeauthor{akhound-sadeghIteratedDenoisingEnergy2024a}
\yrcite{akhound-sadeghIteratedDenoisingEnergy2024a}, which are based on the Real
NVP coupling flow architecture \cite{dinhDensityEstimationUsing2017}. We
repeated experiments with FAB based on the more expressive neural splines
coupling flows \cite{durkanNeuralSplineFlows2019a}, reported as FAB (NEURAL
SPLINES) in Table~\ref{tab:SI_GMM_results}. The same architecture was used for
our TA-BG experiments.

As one can see both from Figure~\ref{SI:fig_GMM} and
Table~\ref{tab:SI_GMM_results}, TA-BG and FAB obtain strong results on the 2D
GMM system, outperforming or matching the considered diffusion-based baselines.
The results of TA-BG and FAB are practically identical and likely close to the
optimal solution possible with the given architecture.

We note that the 2D GMM system is a very simple target, thus the inclusion here
mostly serves illustrative purposes to show that our methodology can also be
applied to more traditional sampling tasks. Since the informative value of a
systematic comparison on this system is limited, we did not tune the results of
TA-BG and FAB with respect to the number of target energy evaluations.

TA-BG can, of course, also be applied to other sampling tasks considered in
related literature. However, we purposely set the focus of this work on
molecular systems, where the potential energy is more correlated and complex
compared to many of the toy problems currently considered in the variational
sampling literature. Symmetric target densities with identical particles, such
as DW-4, LJ-13, or LJ-55 \cite{akhound-sadeghIteratedDenoisingEnergy2024a},
should ideally be handled with an equivariant flow architecture, thus we did not
consider them here.

\begin{table*}[h]
\caption{Comparison of metrics on the 2D GMM system. The best-performing method for each metric is highlighted in bold. For our 
experiments (FAB (NEURAL SPLINES) and TA-BG (NEURAL SPLINES)), we specify the mean
and standard deviation of the metrics across four independent runs, the rest is reported with 
the mean and standard deviation over three independent runs.}
\label{tab:SI_GMM_results}
\vskip 0.15in
\begin{center}
\begin{small}
\begin{sc}
\begin{tabular}{ccccc}
\toprule
METHOD & NLL $\downarrow$ & ESS $\uparrow$ \\
\midrule
FAB (REAL NVP) & \num{7.14(01)} & \SI{65.3(1.7)}{\percent} \\
FAB (neural splines) & \textbf{\num{6.92(0.00)e0}} & \textbf{\SI{97.09(0.58)}{\percent}} \\
TA-BG (neural splines) & \textbf{\num{6.91(0.01)e0}} & \textbf{\SI{96.89(0.48)}{\percent}} \\
\midrule
PIS & \num{7.72(03)} & \SI{29.5(1.8)}{\percent} \\
DDS & \num{7.43(46)} & \SI{68.7(20.8)}{\percent} \\
iDEM & \textbf{\num{6.96(07)}} & \SI{73.4(9.2)}{\percent} \\
\bottomrule
\end{tabular}
\end{sc}
\end{small}
\end{center}
\vskip -0.1in
\end{table*}

\subsection{Details}
Here, we specify hyperparameters and other details for TA-BG and FAB applied to
the GMM system. We only specify hyperparameters that differ from the ones used
in our main experiments on the molecular systems.

Contrary to the experiments on the molecular systems, we did not use a
regularized energy function for the GMM system. Also, we did not clip importance
weights when calculating the ESS. Following previous work, the ESS was
calculated using $1000$ samples from the model, the NLL was calculated using
$1000$ ground truth samples.

\paragraph{Architecture} For all experiments on the GMM system, we used
monotonically increasing rational-quadratic splines with $K=16$ bins in the
range $[-50,50]$. For the parameter networks, we used fully connected neural
networks with hidden dimensions $[120,120]$ and ReLU activation functions. We
used $13$ coupling layers, swapping the two dimensions after each coupling. As
the latent distribution $q_Z$, we used a 2D normal distribution with $\mu=0$ and
$\sigma=10.0$, truncated to the range $[-50,50]$.

\paragraph{FAB Hyperparameters}
We used a learning rate of \num{1e-5}, a batch size of $8192$, and \num{50000}
gradient descent steps. We scaled the coordinates from $[-50,50]$ to $[-5,5]$ to
perform AIS. Contrary to the experiments on the molecular systems, we set both
the number of intermediate AIS distributions and the number of inner HMC steps
to one. We further used $0.05$ as the initial HMC step size.

\paragraph{TA-BG Hyperparameters}
Reverse KLD pre-training was performed at $T=30.0$, using a learning rate of
\num{1e-4}, a batch size of $128$, and \num{50000} gradient descent steps.
Contrary to the reverse KLD experiments on the molecular systems, we here did
not remove the largest energy values in the loss contributions of each batch.
We also did not use a learning rate scheduler.

For the annealing, we used a learning rate of \num{5e-5}, a batch size of
$8192$, and no learning rate scheduler. We used the following annealing
iterations, following a geometric temperature schedule:

\begin{align}
    \underbrace{\SI{30.0}{\kelvin}}_{T_1} \rightarrow \underbrace{\SI{18.45}{\kelvin}}_{T_2} \rightarrow \SI{11.35}{\kelvin} \rightarrow \SI{6.98}{\kelvin}
    \rightarrow \SI{4.30}{\kelvin} \rightarrow \SI{2.64}{\kelvin} \rightarrow \SI{1.63}{\kelvin} \rightarrow \underbrace{\SI{1.0}{\kelvin}}_{T_{K-1}} \rightarrow \underbrace{\SI{1.0}{\kelvin}}_{T_K=T_\text{target}}
\end{align}

For each annealing iteration, we draw \num{2000000} samples, resample to
\num{2000000} samples using the importance weights (without clipping), and
perform \num{20000} forward KLD gradient descent steps.

\clearpage
\section{Additional Figures}
\subsection{All Systems (Reweighted)}

\begin{figure*}[h]
\begin{center}
\vskip 0.2in
\centerline{\includegraphics{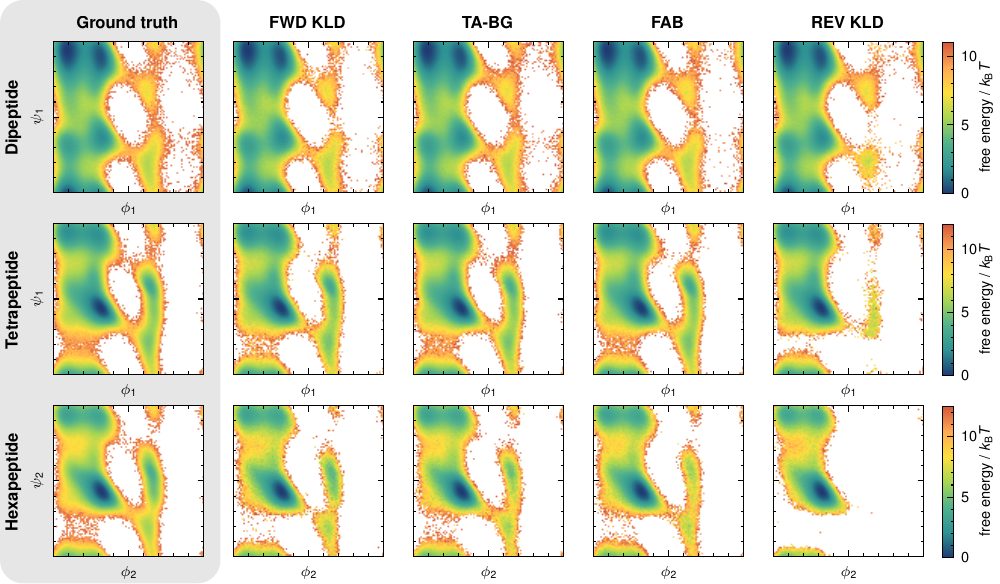}}
\caption{Reweighted version of Figure~\ref{main:fig_ram_comparison}. Comparison
of the free energy $ F=-k_\text{B} T \ln p(\phi_i, \psi_i) $ of selected
dihedral angles (Ramachandran plots) at \SI{300}{\kelvin}, reweighted to
\SI{300}{\kelvin}.}
\label{SI:fig_ram_comparison_reweighted}
\end{center}
\vskip -0.2in
\end{figure*}

\clearpage
\subsection{Alanine Tetrapeptide}

\begin{figure*}[h]
\begin{center}
\vskip 0.2in \centerline{\includegraphics{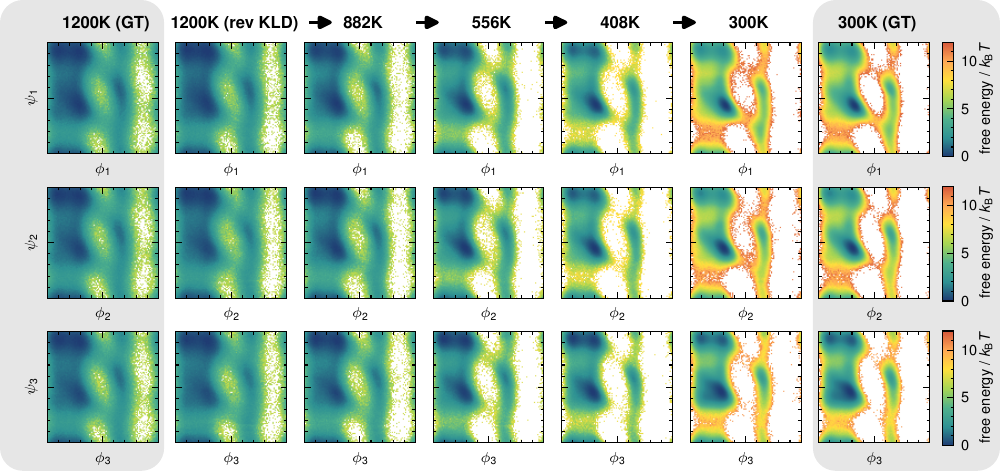}}
\caption{Visualization of the iterative annealing process for the tetrapeptide,
showing the free energy $ F=-k_\text{B} T \ln p(\phi_i, \psi_i) $ of backbone
dihedral angles (Ramachandran plots) in each iteration. Note that not all
annealing iterations are shown. We used \num{1e7} samples for the Ramachandran
plots at \SI{300}{\kelvin} and \num{1e6} samples for the rest.}
\label{SI:tetra_annealing}
\end{center}
\vskip -0.2in
\end{figure*}

\begin{figure*}[h]
\begin{center}
\vskip 0.2in
\centerline{\includegraphics{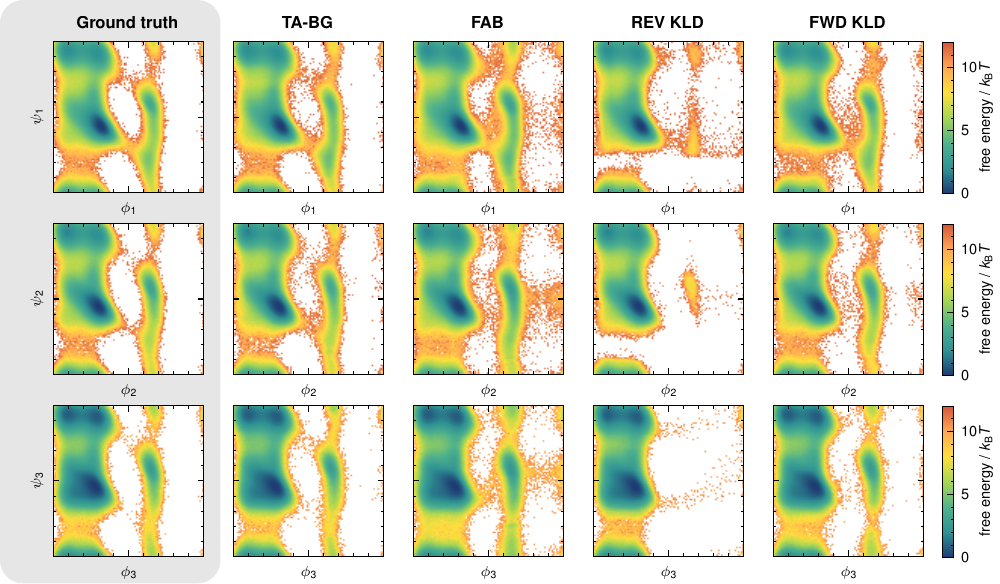}}
\caption{Comparison of the free energy $ F=-k_\text{B} T
\ln p(\phi_i, \psi_i) $ of the backbone dihedral angles (Ramachandran plots) of the
tetrapeptide at \SI{300}{\kelvin}.}
\label{SI:tetra_all_ramachandrans}
\end{center}
\vskip -0.2in
\end{figure*}

\begin{figure*}[h]
\begin{center}
\vskip 0.2in
\centerline{\includegraphics{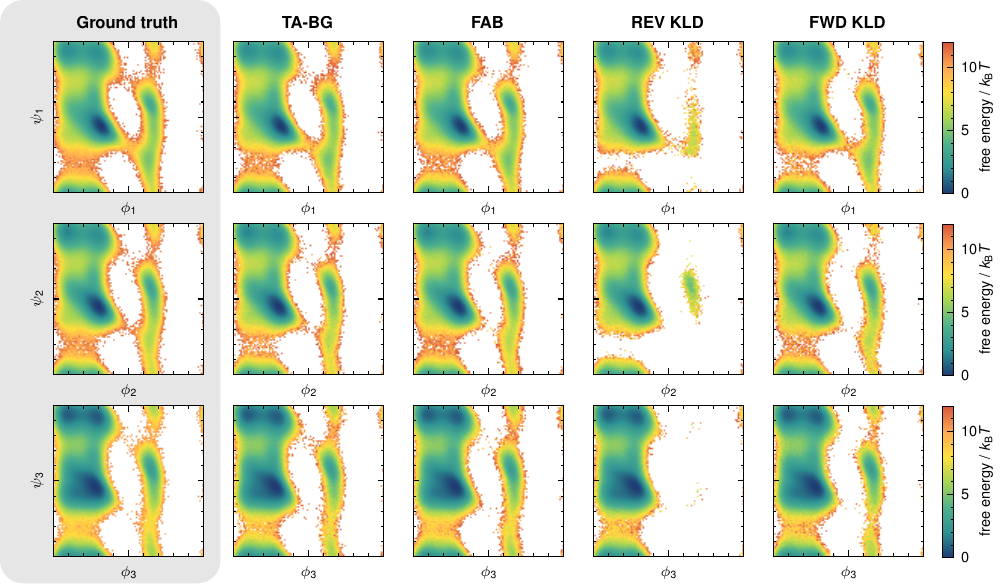}}
\caption{Reweighted version of Figure~\ref{SI:tetra_all_ramachandrans}.
Comparison of the free energy $ F=-k_\text{B} T
\ln p(\phi_i, \psi_i) $ of the backbone dihedral angles (Ramachandran plots) of the
tetrapeptide at \SI{300}{\kelvin}, reweighted to \SI{300}{\kelvin}.}
\label{SI:tetra_all_ramachandrans_reweighted}
\end{center}
\vskip -0.2in
\end{figure*}

\clearpage
\subsection{Alanine Hexapeptide}

\begin{figure*}[h]
\begin{center}
\vskip 0.2in \centerline{\includegraphics{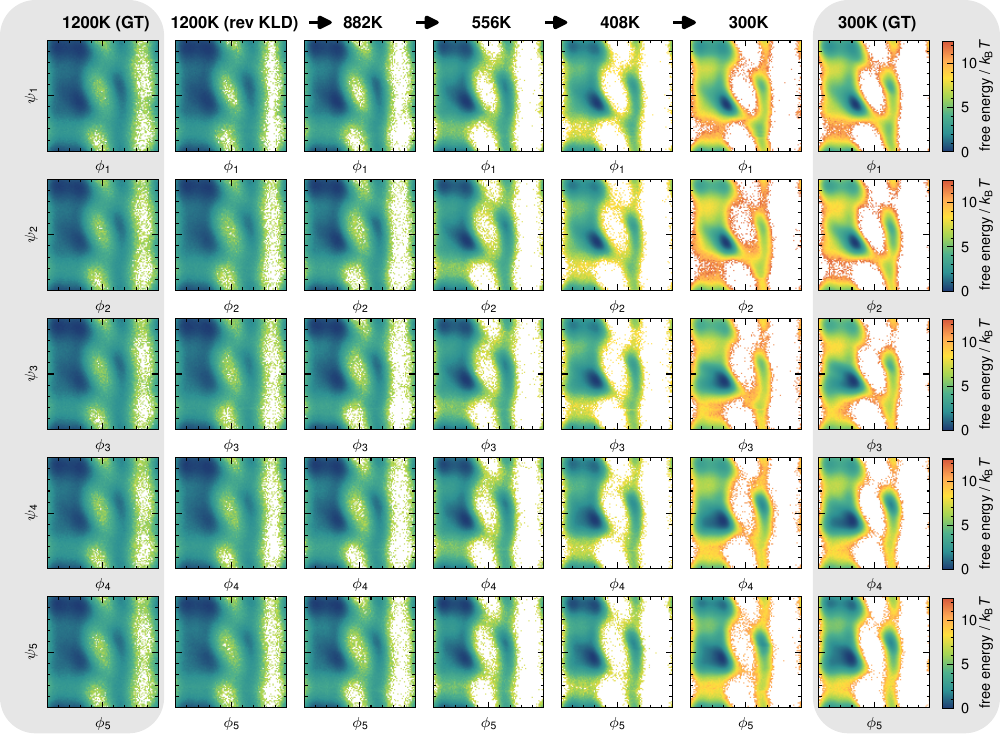}}
\caption{Visualization of the iterative annealing process for the hexapeptide,
showing the free energy $ F=-k_\text{B} T \ln p(\phi_i, \psi_i) $ of backbone
dihedral angles (Ramachandran plots) in each iteration. Note that not all
annealing iterations are shown. We used \num{1e7} samples for the Ramachandran
plots at \SI{300}{\kelvin} and \num{1e6} samples for the rest.}
\label{SI:hexa_annealing}
\end{center}
\vskip -0.2in
\end{figure*}

\begin{figure*}[t]
\begin{center}
\vskip 0.2in
\centerline{\includegraphics{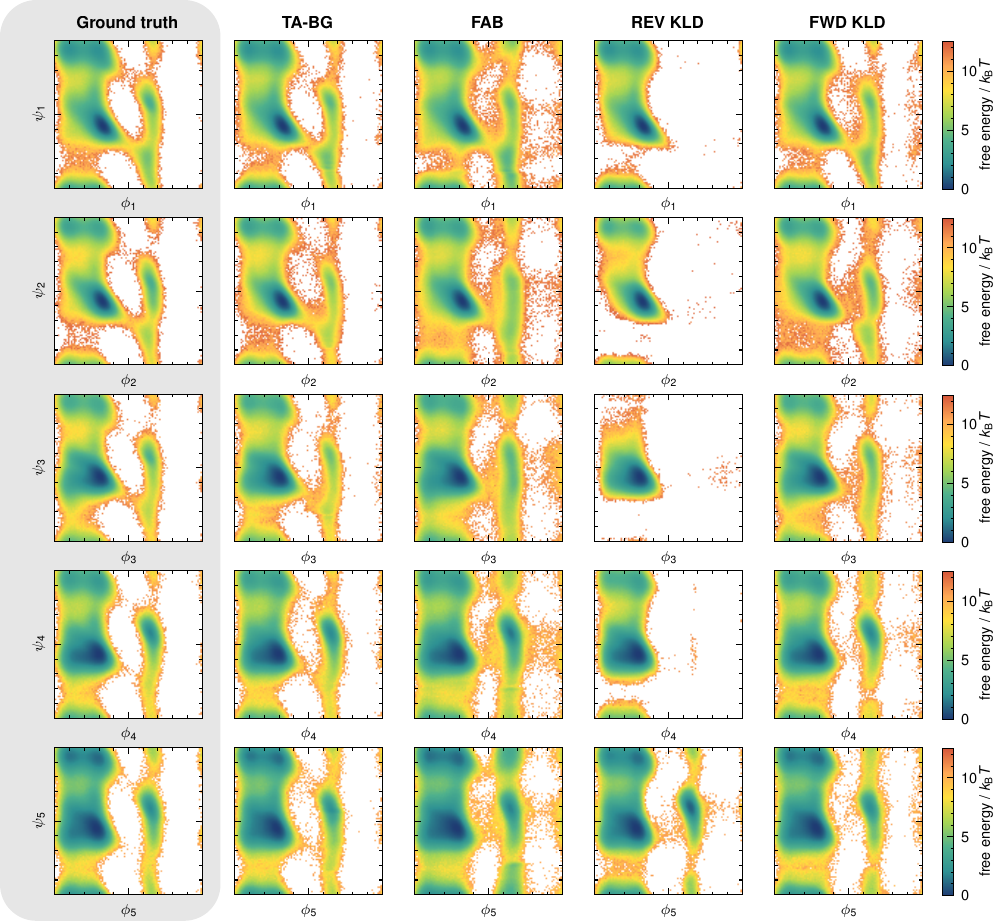}}
\caption{Comparison of the free energy $ F=-k_\text{B} T \ln p(\phi_i, \psi_i) $
of the backbone dihedral angles (Ramachandran plots) of the hexapeptide at \SI{300}{\kelvin}.}
\label{SI:hexa_all_ramachandrans}
\end{center}
\vskip -0.2in
\end{figure*}

\begin{figure*}[h]
\begin{center}
\vskip 0.2in
\centerline{\includegraphics{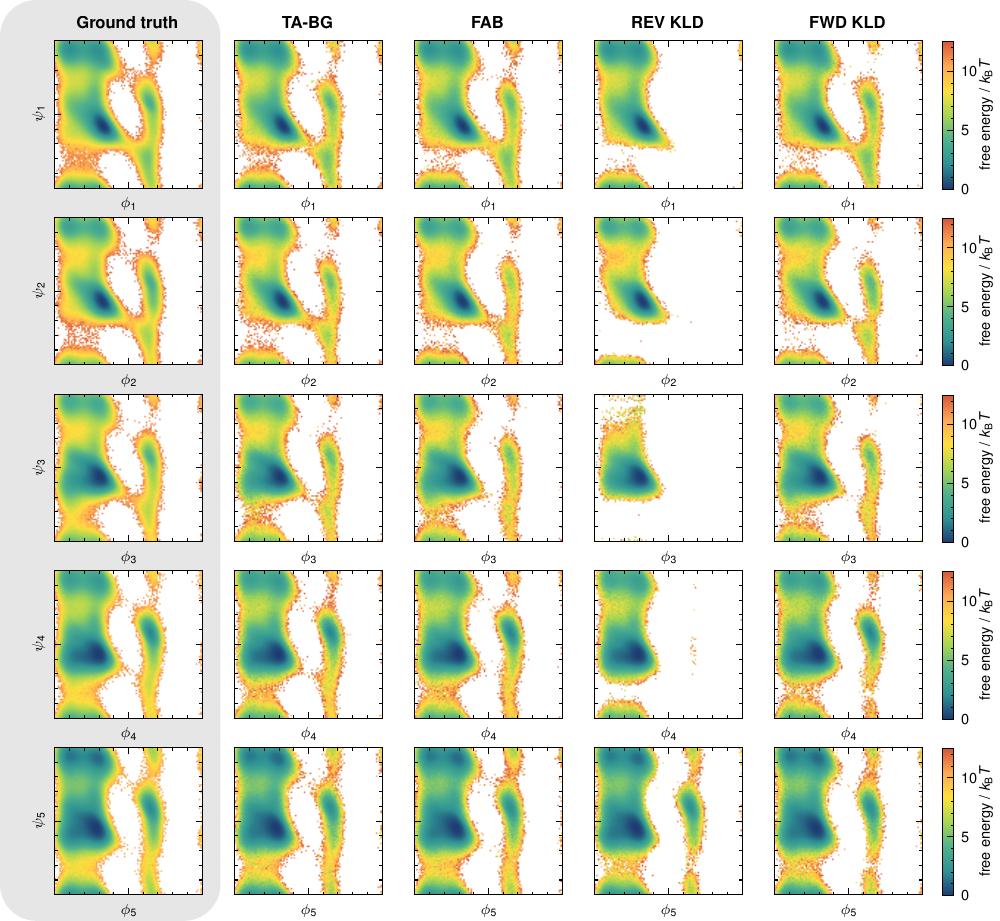}}
\caption{Reweighted version of Figure~\ref{SI:hexa_all_ramachandrans}.
Comparison of the free energy $ F=-k_\text{B} T \ln p(\phi_i, \psi_i) $ of the
backbone dihedral angles (Ramachandran plots) of the hexapeptide at \SI{300}{\kelvin}, reweighted to
\SI{300}{\kelvin}.}
\label{SI:hexa_all_ramachandrans_reweighted}
\end{center}
\vskip -0.2in
\end{figure*}

\end{document}